\documentclass[journal]{IEEEtran}
\usepackage[utf8]{inputenc}
\usepackage{csquotes}
\usepackage{hyperref}
\usepackage{graphicx}
\usepackage{multicol, multirow, bigstrut, rotating}
\usepackage{dirtytalk}
\ifCLASSOPTIONcompsoc
  \usepackage[caption=false,font=normalsize,labelfont=sf,textfont=sf]{subfig}
\else
  \usepackage[caption=false,font=footnotesize]{subfig}
\fi

\usepackage[bibstyle=ieee,
            citestyle=numeric-comp,
            minnames=1, maxnames=3,
            backend=biber,
            doi=false,isbn=false,
            url=false]{biblatex}
\AtEveryBibitem{\clearfield{month}}
\AtEveryCitekey{\clearfield{month}}

\usepackage{todonotes, comment, verbatim}
\usepackage{xcolor, soul}
\usepackage{enumitem}
\setlist{nolistsep}

\newcommand{\etal}{\textit{et al}.}
\newcommand{\eg}{e.g.,}
\newcommand{\ie}{i.e.,}

\newcommand{\githubLGKFgS}{\href{https://github.com/gait-tech/gt.papers/tree/master/+ckf2019}{https://git.io/Je9VV}}

\newcommand{\CKFcIMU}{\emph{CKF-3I}}
\newcommand{\LGKFeSegcIMU}{\emph{L5S-3I}}
\newcommand{\LGKFgSeg}{\emph{L7S}}
\newcommand{\LGKFgSegcIMU}{\emph{L7S-3I}}
\newcommand{\LGKFgSegbIMU}{\emph{L7S-2I}}

\usepackage{mdframed}
\usepackage{empheq}
\iffalse
    \global\mdfdefinestyle{mdchange}{%
    backgroundcolor=yellow, linewidth=0pt,%
    leftmargin=0pt,rightmargin=0pt,
    skipabove=0,skipbelow=0,
    innerleftmargin=0pt,innerrightmargin=0pt,
    innertopmargin=0pt,innerbottommargin=0pt
    }
    \DeclareRobustCommand{\changed}[1]{\hl{#1}}
    
\else
    \global\mdfdefinestyle{mdchange}{%
    linewidth=0pt,%
    leftmargin=0pt,rightmargin=0pt,
    skipabove=0,skipbelow=0,
    innerleftmargin=0pt,innerrightmargin=0pt,
    innertopmargin=0pt,innerbottommargin=0pt
    }
    \sethlcolor{white}
    \DeclareRobustCommand{\changed}[1]{\hl{#1}}
    
\fi

\usepackage{amsmath, amssymb, bm, units, arydshln}
\usepackage{amsmath, amssymb, bm, units}
\usepackage{mathrsfs}
\usepackage{nicefrac, xfrac, arydshln}
\usepackage{mathtools}

\newcommand{\cs}[1]{#1}
\newcommand{\bsraw}[5]{\prescript{#1}{#2}{#3}^{#4}_{#5}}
\newcommand{\bsUR}[5]{ 
	\ifthenelse{\equal{#4}{} \OR \equal{#2}{}}{
		\bsraw{\cs{#1}}{}{#3}{\cs{#2}#4}{#5}
	}{
		( \bsraw{\cs{#1}}{}{#3}{\cs{#2}}{#5} )^{#4}
	}
}

\newcommand{\bs}[5]{
	\ifthenelse{\equal{#1}{W}}{
		\bsUR{}{#2}{#3}{#4}{#5}
	}{
		\bsUR{#1}{#2}{#3}{#4}{#5}
	}
}

\newcommand{\bvraww}[5]{\prescript{\cs{#1}}{#2}{\vec{#3}}^{#4}_{#5}}

\newcommand{\bvUR}[5]{
	\ifthenelse{\equal{#4}{} \OR \equal{#2}{}}{
		\bvraww{#1}{}{#3}{\cs{#2}#4}{#5}
	}{
		( \bvraww{#1}{}{#3}{\cs{#2}}{#5} )^{#4}
	}
}
\newcommand{\bv}[5]{
	\ifthenelse{\equal{#1}{W}}{
		\bvUR{}{#2}{#3}{#4}{#5}
	}{
		\bvUR{#1}{#2}{#3}{#4}{#5}
	}
}
\newcommand{\bvmeas}[5]{
	\bv{#1}{#2}{\breve{#3}}{#4}{#5}
}

\newcommand{\R}{\mathbb{R}}
\newcommand{\N}{\mathcal{N}}

\newcommand{\mat}[1]{\mathbf{\bm{#1}}}
\renewcommand{\vec}[1]{\mathbf{\bm{#1}}}
\newcommand{\smallpct}{\scalebox{0.5}[1.0]{\%}}
\newcommand{\smallneg}{\scalebox{0.5}[1.0]{$-$}}
\newcommand{\smallpm}{\scalebox{0.5}[1.0]{$\pm$}}
\newcommand\LG[1]{#1} 
\newcommand\LA[1]{\mathfrak{#1}} 
\newcommand\Lvee[2]{\left[#2\right]^{\vee}_{\LG{#1}}}
\newcommand\Lhat[2]{\left[#2\right]^{\wedge}_{\LG{#1}}}
\newcommand\Lodot[2]{\left[#2\right]^{\odot}_{\LG{#1}}}

\newcommand\LJac[2]{\Phi_{\LG{#1}}\left(#2\right)}
\newcommand\Lvectran[2]{\exp\left(\Lhat{#1}{#2}\right)}
\newcommand\Ltranvec[2]{\Lvee{#1}{\log\left( #2 \right)}}

\newcommand\LveeSM[2]{[#2]^{\vee}_{\LG{#1}}}
\newcommand\LhatSM[2]{[#2]^{\wedge}_{\LG{#1}}}

\newcommand\LAdSM[2]{\text{Ad}_{\LG{#1}}(#2)}
\newcommand\LadSM[2]{\text{ad}_{\LG{#1}}(#2)}
\newcommand\LJacSM[2]{\Phi_{\LG{#1}}(#2)}
\newcommand\LvectranSM[2]{\exp(\LhatSM{#1}{#2})}
\newcommand\LtranvecSM[2]{\LveeSM{#1}{\log(#2)}}

\newcommand{\diag}{\text{diag}}

\newcommand\bovermat[2]{%
	\makebox[0pt][l]{$\smash{\overbrace{\phantom{%
					\begin{matrix}#2\end{matrix}}}^{\text{#1}}}$}#2}

\newcommand{\quatUR}[5]{
	\ifthenelse{\equal{#4}{} \OR \equal{#2}{}}{
		\prescript{\cs{#1}}{}{\bm{#3}}^{\cs{#2}#4}_{#5}
	}{
		(\prescript{\cs{#1}}{}{\bm{#3}}^{\cs{#2}}_{#5})^{#4}
	}
}
\newcommand{\quatURR}[5]{
	\ifthenelse{\equal{#1}{W}}{
		\quatUR{}{#2}{#3}{#4}{#5}
	}{
		\quatUR{#1}{#2}{#3}{#4}{#5}
	}
}

\newcommand{\quat}[4]{
	\quatURR{#1}{#2}{q}{#3}{#4}
}

\newcommand{\kfsp}[2]{\boldsymbol{\hat{#1}}^{-}_{#2}} 
\newcommand{\kfsm}[2]{\boldsymbol{\hat{#1}}^{+}_{#2}} 
\newcommand{\kfsc}[2]{\boldsymbol{\tilde{#1}}^{+}_{#2}} 
\newcommand{\kfcp}[2]{\mat{#1}^{-}_{#2}} 
\newcommand{\kfcm}[2]{\mat{#1}^{+}_{#2}} 
\newcommand{\kfbvp}[5]{\bv{#1}{#2\text{--}}{\hat{#3}}{#4}{#5}}

\newcommand{\kfbvc}[5]{\bv{#1}{#2\text{+}}{\tilde{#3}}{#4}{#5}}

\iftrue
    \newcommand{\kfsep}[2]{\boldsymbol{\hat{#1}}^{\boldsymbol{\epsilon}-}_{#2}} 
    \newcommand{\kfsem}[2]{\boldsymbol{\hat{#1}}^{\boldsymbol{\epsilon}+}_{#2}} 
    \newcommand{\kfsec}[2]{\boldsymbol{\tilde{#1}}^{\boldsymbol{\epsilon}+}_{#2}} 

\else
    \newcommand{\kfsep}[2]{\boldsymbol{\hat{#1}}^{-}_{#2} \LvectranSM{G}{\epsilon}} 
    \newcommand{\kfsem}[2]{\boldsymbol{\hat{#1}}^{+}_{#2} \LvectranSM{G}{\epsilon}} 
    \newcommand{\kfsec}[2]{\boldsymbol{\tilde{#1}}^{+}_{#2} \LvectranSM{G}{\epsilon}} 

\fi

\newcommand{\dt}{\Delta t}

\newcommand\LKFFunc[4]{#1^{#2}_{#3}(#4)}

\newcommand\LtranvecKFFunc[5]{\LtranvecSM{#1}{\LKFFunc{#2}{#3}{#4}{#5}}}

\usepackage{amssymb}
\usepackage{graphicx}

\soulregister\cite7
\soulregister\ref7
\soulregister\pageref7
\soulregister\eqref7
\soulregister\etal7

\title{Estimating Lower Body Kinematics using a\\
Lie Group Constrained Extended Kalman Filter\\
and Reduced IMU Count
}
\author{Luke Sy \emph{Student Member, IEEE}, 
        Nigel H. Lovell \emph{Fellow, IEEE}, 
        Stephen J. Redmond \emph{Senior Member, IEEE}}
\date{5 May 2020}
\IEEEaftertitletext{\vspace{-2\baselineskip}}

\begin{document}
	\maketitle
	\bstctlcite{IEEEexample:BSTcontrol}
		
	\begin{abstract}
	    \emph{Goal:} This paper presents an algorithm for estimating pelvis, thigh, shank, and foot kinematics during walking using only two or three wearable inertial sensors.
	    \emph{Methods:} The algorithm makes novel use of a Lie-group-based extended Kalman filter. 
    	    The algorithm iterates through the prediction (kinematic equation), measurement (pelvis position pseudo-measurements, zero-velocity update, and flat-floor assumption), and constraint update (hinged knee and ankle joints, constant leg lengths).
	    \emph{Results: } The inertial motion capture algorithm was extensively evaluated on two datasets showing its performance against two standard benchmark approaches in optical motion capture (\ie{} plug-in gait (commonly used in gait analysis) and a kinematic fit (commonly used in animation, robotics, and musculoskeleton simulation)),
            giving insight into the similarity and differences between the said approaches used in different application areas.
        The overall mean body segment position (relative to mid-pelvis origin) and orientation error magnitude of our algorithm ($n=14$ participants) for free walking was $5.93 \pm 1.33$ cm and $13.43 \pm 1.89^\circ$ when using three IMUs placed on the feet and pelvis, 
            and $6.35 \pm 1.20$ cm and $12.71 \pm 1.60^\circ$ when using only two IMUs placed on the feet.
	    \emph{Conclusion: } The algorithm was able to track the joint angles in the sagittal plane for straight walking well, but requires improvement for unscripted movements (\eg{} turning around, side steps), especially for dynamic movements or when considering clinical applications.
	    \emph{Significance: } This work has brought us closer to comprehensive remote gait monitoring using IMUs on the shoes.
            The low computational cost also suggests that it can be used in real-time with gait assistive devices.
	\end{abstract}

    
\section{Introduction}
The tracking of human body movement has not only fascinated researchers for years,
    but has also recently found application in robotics, virtual reality, animation, and healthcare (\eg{} gait analysis).
Human pose (\eg{} body joint kinematics) is typically captured within a confined space using an optical motion capture (OMC) system capable of estimating position up to millimeter accuracy (assuming it is well calibrated).
Many commercial OMC systems use passive or active surface markers attached to the skin above bony landmarks to estimate the kinematics of the skeleton.
In gait analysis, the skeletal kinematics are usually estimated using one of two approaches: direct kinematics and inverse kinematics.
Direct kinematic analysis involves estimating pose (\ie{} position and orientation of body segments) directly from the markers (\eg{} Vicon's Plug-in Gait) \cite{Kadaba1990}.
    It is typically used in gait analysis.
    \changed{
    However, correct and systematic marker placement are extremely important to obtain accurate and consistent pose reconstruction \cite{Leardini2005} (\ie{} a trained personnel is needed for marker placement).}
Inverse kinematics estimates the best skeletal pose by optimising the pose of a linked-segment model of the skeleton to best match the captured OMC marker data.
    It is typically used in robotics, animation, and in musculoskeletal modelling software (\eg{} OpenSim) \cite{Delp2007a}.
    This approach may also take advantage of simple joint constraints (\eg{} hinged knee joint) which can ultimately reduce inter-trial variability and, possibly, soft tissue artifacts, but at the cost of an inability to capture certain pathological conditions where these constraints are not respected \cite{Kainz2017a}.
There is no definitive/universal kinematic model of the body in the literature \cite{Leardini2017}.
\changed{The estimated pose from both approaches can be very similar, specially when the markers are placed perfectly and if the subject is healthy.
Nevertheless, the fact remains that} each approach has limitations, and the most appropriate model may ultimately depend on the application.

The miniaturization and low cost of inertial measurements units (IMUs) has enabled the development of inertial motion capture (IMC) systems.
IMC systems can operate independently from any fixed external sensor (\eg{} cameras).
Compared to OMC systems, they are immune to occlusion and lighting issues making them suitable for prolonged use outside of the laboratory.
    However, the lack of an external position reference can lead to positional drift (\ie{} root position of body in the global frame becomes lost).
Commercial IMCs typically attach one sensor per body segment (OSPS) \cite{Roetenberg2009},
which may be considered too cumbersome and expensive for routine daily use due to the number of sensor units required.
    The orientation of each body segment is tracked by the attached IMU using an orientation estimation algorithm (\eg{} \cite{DelRosario2016a, DelRosario2018}), 
        which is then combined via a linked kinematic chain to estimate body pose, similar to OMC inverse kinematic models, usually rooted at the pelvis.

Recent advances in IMC algorithms have made possible motion capture using a reduced-sensor-count (RSC) configuration, 
where IMUs are placed on a subset of body segments.
Such configurations can improve user comfort while also reducing setup time and system cost.
However, utilizing fewer sensors inherently reduces the amount of kinematic information available; 
this information must be inferred by enforcing mechanical joint constraints \cite{syckf2020}, making dynamic balance assumptions, or using additional sensors (\eg{} cameras or distance measurement \cite{Vlasic2007, syckfdist2020}).
    Amongst additional sensor approaches, video-inertial systems are the most common, where IMU measurements help resolve orientation ambiguity for OMC systems \cite{Fakult2013, VonMarcard2016, Trumble2017, Gilbert2019}.
Developing a self-contained and comfortable IMC system for routine daily use may facilitate interactive rehabilitation (\eg{} provide real-time feedback to improve walking stability \cite{Llorens2015} or reduce joint loading \cite{Shull2010}),
    and possibly track/study the progression of movement disorder to enable predictive diagnostics.
IMC algorithms can be classified into two main approaches: data-driven and model-based.

Data-driven approaches statistically infer the kinematics of uninstrumented segments by comparing sensor measurement patterns (or derivatives of it) to an existing motion database (DB) (\eg{} nearest-neighbor search using either one or multiple time steps of past movements \cite{Tautges2011, Wouda2016}),
    or train some model using the data (\eg{} shallow and deep neural network (NN) \cite{Wouda2016, Wouda2019}, bi-directional recurrent NN which take into account temporal information \cite{Huang2018a}). 
    The assumption is that the kinematics of body segments without sensors are well correlated with the kinematics of body segments with sensors.
Indeed, such is the case for the movements of healthy subjects, which could be why data-driven approaches have been shown effective in reconstructing realistic motions for animation-related applications \cite{Tautges2011, Wouda2016, Huang2018a, Wouda2019}.
However, the pose reconstruction for these approaches naturally have a bias toward motions represented in the training DB, inherently limiting their use to novel movements not contained in the training DB (\eg{} pathological gait monitoring).
    
Model-based approaches infer the kinematics of uninstrumented segments by leveraging kinematic and biomechanical models, 
similar to OMC inverse kinematic approaches which model the human body as linked rigid body segments.
Early works started in 2D tracking (\eg{} linear regression model \cite{Salarian2013}, inverse kinematic of legs in the sagittal plane \cite{Hu2015}), which can have difficulty tracking body movement during some activities of daily living (ADLs), such as side or diagonal steps.
Recent literature has shown that sparse motion capture is also possible in 3D (\eg{} window-based optimization on full body segments linked by 24 ball and socket joints \cite{Marcard2017}).
In our recent work, we tracked five body segments (\ie{} the pelvis, thigh, and shanks) using IMUs at the pelvis and ankles using a constrained Kalman filter (CKF) where orientation was represented using quaternions \cite{syckf2020}. 
    Building on prior work on state estimation using a Lie group representation (\cite{Wang2006, Barfoot2014, Bourmaud2013} for propagating uncertainty, \cite{Joukov2017} for IMC systems under OSPS configuration), we further extended their work by representing and tracking pose using Lie groups, specifically the special Euclidean group,  $\LG{SE(3)}$ \cite{sylgcekf2020}.
    Tracking orientation using Lie groups is arguably more elegant,
        as it does not require additional constraints, such as those required by rotation matrix or quaternion representations (\eg{} constraints $\mat{R}^T\mat{R} = \mat{I}$ or $||\quat{}{}{}{}||=1$) \cite{Barfoot2014},
        while providing significant improvements over an Euler angle representation in near-gimbal-lock poses \cite{Joukov2017}.

\subsection{Novelty}
This paper describes a novel 3D lower body pose estimator that uses a constrained Lie group Kalman filter using RSC configuration of IMUs.
It builds on prior work \cite{sylgcekf2020} but instead tracks all seven major lower body segments, instead of five, using only two or three IMUs.
In our prior work \cite{sylgcekf2020}, the orientation of the thigh is inferred from the tracked pelvis and shank poses.
In this work, the orientations of the uninstrumented segments (\ie{} thighs and shanks) are inferred from the poses of the pelvis and feet, tracked by the CKF.
As this algorithm achieves a low computation cost compared to data-driven and optimization-based algorithms, it can be used in real-time applications.
This design was motivated by the need to develop a gait assessment tool using as few a number of sensors as possible, ergonomically-placed for comfort, to facilitate long-term monitoring of lower body movement.
We believe having the IMUs on/in the shoes is more convenient and comfortable than attaching them to the ankles or shanks (\eg{} \cite{sylgcekf2020, Marcard2017}),
and allows for more accurate step detection performance. 
Lastly, the algorithm was extensively tested on two types of OMC benchmark, direct kinematics (\ie{} Plug-in Gait) commonly used in gait analysis, and an inverse kinematics model, commonly used in musculoskeleton modelling and robotics.

\section{Algorithm description}
\begin{figure}[htbp]
    \centering
    \begin{minipage}{0.7\linewidth}
    \includegraphics[width=\linewidth]{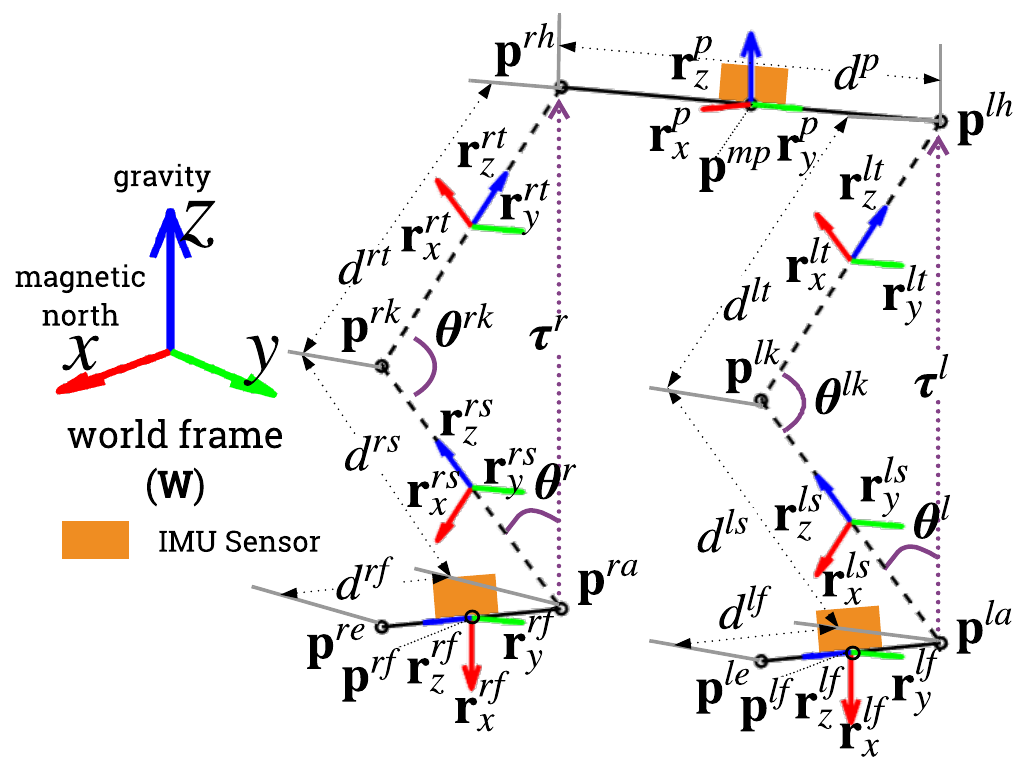}
    \end{minipage}
    \begin{minipage}{0.28\linewidth}
        \scriptsize
        \begin{tabular}{|c|c|l|} \hline 
            & Symb. & Desc. \\ \hline
            \multirow{9}[1]{0.15cm}{\begin{sideways}Points/Joints\end{sideways}} & $mp$ & mid-pelvis \\  
            & $lh$ & left hip \\
            & $rh$ & right hip \\
            & $lk$ & left knee \\
            & $rk$ & right knee \\
            & $la$ & left ankle \\
            & $ra$ & right ankle \\
            & $le$ & left toe \\ 
            & $re$ & right toe \\ \hline
            \multirow{7}[1]{0.15cm}{\begin{sideways}Segments\end{sideways}} & $p$ & pelvis \\
            & $lt$ & left thigh \\
            & $rt$ & right thigh \\
            & $ls$ & left shank \\ 
            & $rs$ & right shank \\
            & $lf$ & left foot \\ 
            & $rf$ & right foot \\ \hline
        \end{tabular}
    \end{minipage} 
    \newline
    \caption{Physical model of the lower body used by the algorithm. The circles denote joint positions, the solid lines denote instrumented body segments (\ie{} pelvis and feet), whilst the dashed lines denote segments without IMUs attached (\ie{} thighs and shanks).}
    \label{fig:body-skeleton}
\end{figure}

\begin{figure}[htbp]
    \centering
    \includegraphics[width=0.45\textwidth]{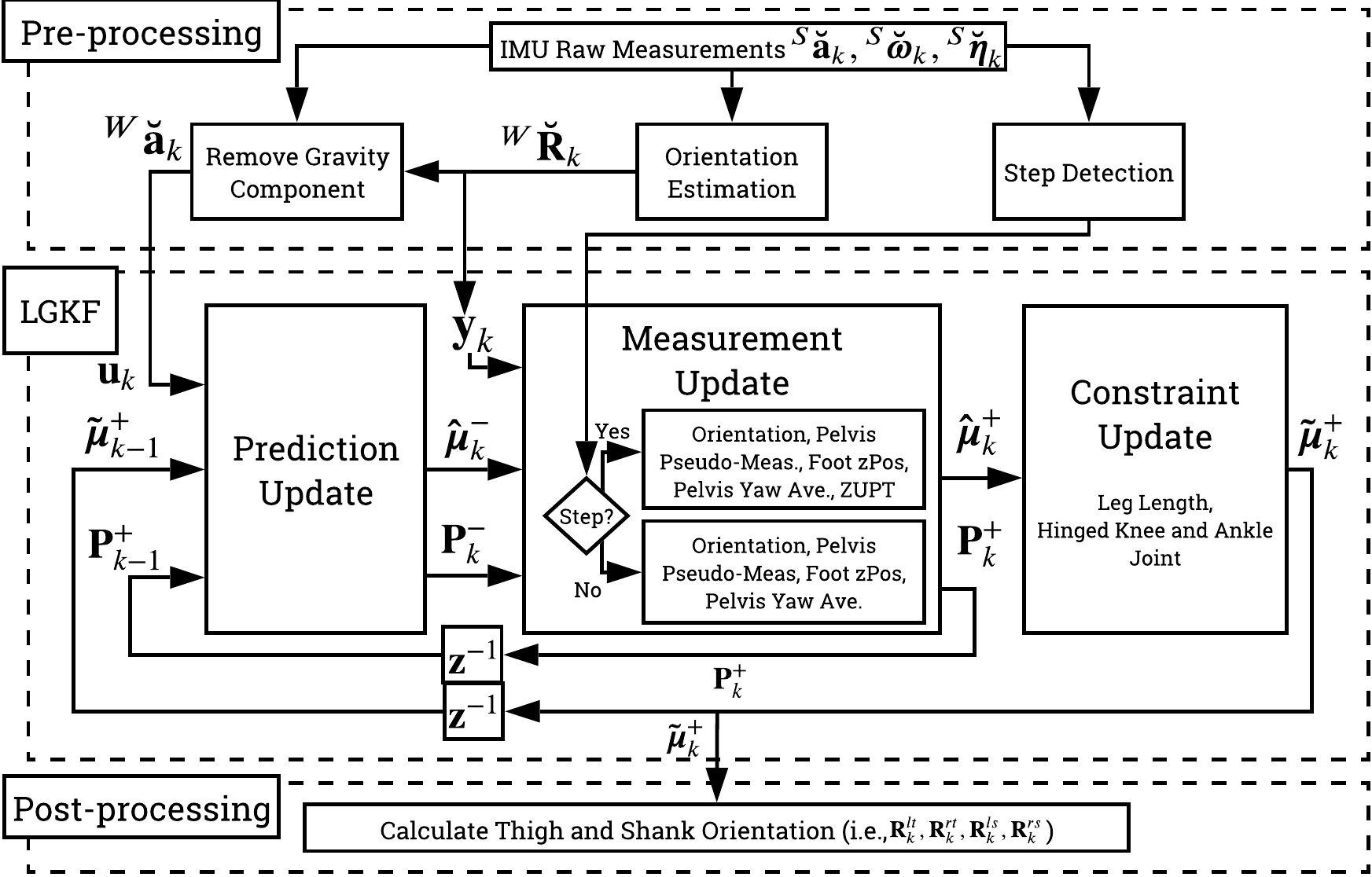}
    \caption{Algorithm overview which consists of pre-processing, LGKF, and post-processing. 
        Pre-processing calculates the body segment orientation, inertial body acceleration, and step detection from raw acceleration, ${}^{S}\vec{\breve{a}}_{k}$, angular velocity, ${}^{S}\vec{\breve{\omega}}_{k}$, and magnetic north heading, ${}^{S}\vec{\breve{\eta}}_{k}$, measured by the IMU in the IMU frame. 
        The LGKF-based state estimation consists of a prediction (kinematic equation), measurement (orientation, pelvis pseudo measurement, $z$-position assumptions, and intermittent zero-velocity update (ZUPT)), and constraint update (maximum leg length, hinged knee and ankle joints). 
        Post-processing calculates the thigh and shank orientations.}
    \label{fig:algo-overview}
\end{figure}

The proposed algorithm, \LGKFgSeg{} (for Lie seven segment), estimates the orientation of the pelvis, thighs, shanks, and feet (\ie{} 7 segments) with respect the world frame, $\cs{W}$, using either two or three IMUs.
    It extends the model and assumptions from our prior work \cite{syckf2020, sylgcekf2020} (\LGKFeSegcIMU{}, \CKFcIMU{}, that aim to estimate the kinematics of five body segments, and places IMUs on the pelvis and shanks).
Two variants of the algorithm are described:
    \LGKFgSegcIMU{} which uses three IMUs attached at the sacrum and feet (Fig. \ref{fig:body-skeleton});
    and \LGKFgSegbIMU{} which uses two IMUs attached at the feet (sacrum IMU pseudo-measurements are estimated by aggregating measurments from the foot IMUs).
Fig. \ref{fig:algo-overview} shows an overview of the proposed algorithm.

\LGKFgSeg{} predicts the position of each foot through double integration of its linear 3D acceleration, as measured by the attached IMUs 
    (after a pre-processing step that resolves these accelerations in the world frame and removes gravitational acceleration).
Orientation is obtained from a third-party orientation estimation algorithm (\eg{} Xsens' algorithm was used in this paper).
To mitigate positional drift due to sensor bias and noise that accumulates in the double integration of acceleration, the pose reconstruction of the instrumented body segments was estimated using the following assumptions:
    (1) the foot 3D velocity and height above the floor are zero whenever a footstep is detected;
    (2) the pelvis $x$, $y$ position is approximately the average of the both feet $x$, $y$ positions (\ie{} a simple balance constraint); and
    (3) the pelvis $z$ position is approximately the length of the unbent leg(s) above the floor.
For \LGKFgSegbIMU{} (\ie{} only two IMUs on the feet and no pelvis IMU), pelvis orientation pseudomeasurement is taken to have zero pitch and roll, and yaw angle equal to the average yaw angle of the two feet. 
Lastly, biomechanical constraints enforce maximum leg length; and hinged knee and ankle joints (one degree of freedom (DOF)).
The pre-processing components of the algorithm are similar as \LGKFeSegcIMU{} \cite{sylgcekf2020},
while the post-processing components are modified to calculate both thigh and shank (instead of just the thigh) orientations from the KF states.

\subsection{Lie group and Lie algebra}
The matrix Lie group $\LG{G}$ is a group of $n \times n$ matrices that is also a smooth manifold.
    It can be used to represent rotation or pose (\eg{} $\LG{SO(3)}$, $\LG{SE(3)}$).
	Group composition and inversion (\ie{} matrix multiplication and inversion) are smooth operations.
The Lie algebra $\LA{g}$ represents a tangent space of a group at the identity element \cite{selig2004lie}.
    The elegance of Lie theory lies in it being able to represent pose using a vector space (\eg{} Lie group $\LG{G}$ is represented by $\LA{g}$) without additional constraints 
        (\eg{} without requiring $\mat{R}^T\mat{R} = \mat{I}$ when using a rotation matrix representation of orientation,
        or $||\quat{}{}{}{}||=1$ when using a quaternion representation of orientation) \cite{stillwell2008naive}.
        	
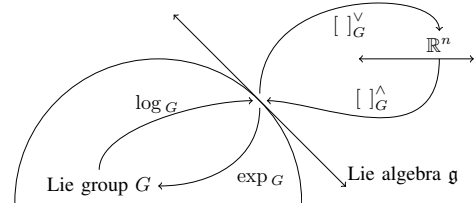
\begin{figure}
    \centering
    \resizebox{0.7\linewidth}{!}{
	\begin{tikzpicture}
		\begin{scope}
		\clip (-2.5,0) rectangle (2.5,2.5);
		\draw (0,0) circle (2.5) node[above left] (A) {Lie group $\LG{G}$};
		\end{scope}
		\draw[<->] (1.77-1.5,1.77+1.5) -- (1.77+1.5,1.77-1.5)
		node[pos=0.5, label={[xshift=2.5cm, yshift=-1.7cm]{Lie algebra $\LA{g}$}}] (B) {};
		\draw[<->] (3.5,2.5) -- (5.5,2.5)
		node[pos=0.7, above] (C) {$\R^n$};
		\draw[->] (A) .. controls +(up:1cm) and +(left:1cm) .. (B)
		node[pos=0.5, above]{$\log{}_{\LG{G}}$};
		\draw[->] (B) .. controls +(down:1cm) and +(right:2cm) .. (A)
		node[pos=0.5, below right]{$\exp{}_{\LG{G}}$};
		\draw[->] (B) .. controls +(up:2cm) and +(up:1cm) .. (C)
		node[pos=0.5, below]{$\Lvee{G}{\:\:}$};
		\draw[->] (C) .. controls +(down:2cm) and +(right:1cm) .. (B)
		node[pos=0.5, above]{$\Lhat{G}{\:\:}$};
	\end{tikzpicture}}
	
    \caption{Mapping between Lie group $\LG{G}$, Lie algebra $\LA{g}$, and $n$-dimensional vector space.
    When $\LG{G} = \LG{SE(3)}$, Lie group $\mat{X} = \mat{T}$ is a $4 \times 4$ transformation matrix representing pose (\ie{} 3D rotation and translation). 
        Similarly, $\vec{v} = \vec{\xi}$ where Lie algebra $\LhatSM{SE(3)}{\vec{\xi}} \in \LA{se(3)}$ and the vector $\vec{\xi} \in \R^n$ with $n=6$.}
    \label{fig:lie-group-algebra-mapping}
\end{figure}

The matrix exponential $\exp{}_{\LG{G}}: \LA{g} \tiny{\to} \LG{G}$ and matrix logarithm $\log{}_{\LG{G}}: \LG{G} \tiny{\to} \LA{g}$ establish a local diffeomorphism between the Lie group $\LG{G}$ and its Lie algebra $\LA{g}$.
    The Lie algebra $\LA{g}$ is a $n \times n$ matrix that can be represented compactly with an $n$-dimensional vector space using the linear isomorphisms (\ie{} one-to-one mappings) 
    $\Lvee{G}{\:\:}: \LA{g} \tiny{\to} \R^n$ and
    $\Lhat{G}{\:\:}: \R^n \tiny{\to} \LA{g}$,
    which map between the compact and matrix representation of the Lie algebra $\LA{g}$.
Fig. \ref{fig:lie-group-algebra-mapping} shows an illustration of the said mappings.
Furthermore, the adjoint operators of a Lie group, $\LAdSM{G}{\mat{X}}$, and its Lie algebra, $\LadSM{G}{\vec{v}}$, where $\mat{X} \in \LG{G}$ and $\Lhat{G}{\vec{v}} \in \LA{g}$, will be used in later sections.
    Multiplying an $n$-dimensional vector representation of a Lie algebra with $\LAdSM{G}{\mat{X}} \in \R^{n \times n}$ (\ie{} the product $\LAdSM{G}{\mat{X}} \vec{v}$) transforms the vector from one coordinate frame to another,
        similar to how rotation matrices transform points from one frame to another.
    A short summary of the operators for Lie groups $\LG{SO(3)}$, $\LG{SE(3)}$, and $\LG{\R^n}$ can be found in \cite{Cesic2016, sylgkf5seg2020}. 
    For a more detailed introduction to Lie groups, refer to \cite{barfoot2017state, stillwell2008naive, Chirikjian2012Book2}.

\subsection{System, measurement, and constraint models}
The system, measurement, and constraint models are presented below
\begin{gather}
    \begin{split}
        \vec{X}_{k} &= f(\vec{X}_{k\smallneg1}, \vec{u}_{k}, \vec{n}_{k}) \\
            &= \vec{X}_{k\smallneg1} \LvectranSM{G}{ \Omega (\vec{X}_{k\smallneg1}, \vec{u}_{k} ) \tiny{+} \vec{n}_{k} } 
    \end{split} \label{eq:pred-update} \\
    \vec{Z}_{k} = h (\vec{X}_{k}) \LvectranSM{G}{\vec{m_{k}}}  ,\quad
    \vec{D}_{k} = c (\vec{X}_{k}) \label{eq:meas-cstr-update}
\end{gather}
where $k$ is the time step.
	$\vec{X}_{k} \in \LG{G}$ is the system state, an element of state Lie group $\LG{G}$.
	$\Omega \left( \vec{X}_{k\smallneg1}, \vec{u}_{k} \right) : \LG{G} \tiny{\to} \R^p$ is a non-linear function which describes how the model acts on the state and input, $\vec{u}_{k}$, where $p$ is the number of dimensions of the compact vector representation for Lie algebra $\LA{g}$.
	$\vec{n}_{k}$ is a zero-mean process noise vector with covariance matrix $\mathcal{Q}$ (\ie{} $\vec{n}_k \sim \N_{\R^p}(\vec{0}_{p \times 1}, \mathcal{Q})$).
	$\vec{Z}_{k} \in \LG{G_m}$ is the system measurement, an element of measurement Lie group $\LG{G_m}$.
	$ h\left(\vec{X}_{k}\right): \LG{G} \tiny{\to} \LG{G_m}$ is the measurement function.
	$\vec{m}_{k}$ is a zero-mean measurement noise vector with covariance matrix $\mathcal{R}_{k}$ (\ie{} $\vec{m}_k \sim \N_{\R^q}(\vec{0}_{q \times 1}, \mathcal{R}_{k})$ where $q$ is the number of dimensions of available measurements).
	$\mat{D}_{k} \in \LG{G_c}$ is the constraint state,
	 an element of constraint Lie group $\LG{G_c}$.
	$ c\left(\vec{X}_{k}\right): \LG{G} \tiny{\to} \LG{G_c}$ is the equality constraint function that state $\vec{X}_{k}$ must satisfy (\ie{} $c\left(\mat{X}_{k}\right) = \mat{D}_k$).
Similar to \cite{Bourmaud2013, Cesic2016}, the state distribution of $\vec{X}_{k}$ is assumed to be a concentrated Gaussian distribution on Lie groups (\ie{} $\vec{X}_{k} = \vec{\mu}_k \Lvectran{G}{\vec{\epsilon}}$, where $\vec{\mu}_k$ is the mean of $\vec{X}_{k}$ and Lie algebra error $\vec{\epsilon} \sim \N_{\R^p}(\vec{0}_{p \times 1}, \mat{P}_k)$) \cite{Wang2006}.

The Lie group state variables $\vec{X}_{k}$ model the position, orientation, and velocity of the three instrumented body segments (\ie{} pelvis and foot) as 
$\vec{X}_k = \diag ( 
		\bv{W}{p}{T}{}{k}, \bv{W}{lf}{T}{}{k}, \bv{W}{lf}{T}{}{k},
		\LvectranSM{\R^9}{[\bv{W}{mp}{v}{T}{k} \: \bv{W}{lf}{v}{T}{k} \: \bv{W}{lf}{v}{T}{k}]^T} 
   	)$ $\in$ $ \LG{G} = \LG{SE(3)}^3 \times \R^{9}$
where $\bv{A}{b}{T}{}{} = \begin{bsmallmatrix} \bv{A}{b}{R}{}{} & \bv{A}{b}{p}{}{} \\ \mat{0}_{1 \times 3} & 1 \end{bsmallmatrix} \in \LG{SE(3)}$ contains orientation $\bv{W}{b}{R}{}{}$ and position $\bv{W}{b}{p}{}{}$ of body segment $b$ relative to frame $A$,
    and $\bv{A}{b}{v}{}{}$ is the velocity of body segment $b$ relative to frame $A$.
If frame $\cs{A}$ is not specified, assume reference to the world frame, $\cs{W}$.
The Lie algebra error is denoted as $\bv{}{}{\epsilon}{}{} = [ \bv{}{p}{\epsilon}{T}{\mat{T}} \:\: \bv{}{lf}{\epsilon}{T}{\mat{T}} \:\: \bv{}{rf}{\epsilon}{T}{\mat{T}} \:\: \bv{}{mp}{\epsilon}{T}{\vec{v}} \:\: \bv{}{lf}{\epsilon}{T}{\vec{v}} \:\: \bv{}{rf}{\epsilon}{T}{\vec{v}} ]^T$ where the first three variables correspond to the Lie group in $\LG{SE(3)}$ while the latter three are for $\R^9$.
$\LveeSM{G}{\cdot}$, $\LvectranSM{G}{\cdot}$, $\LtranvecSM{G}{\cdot}$, $\LAdSM{G}{\vec{X}_k}$, and $\LJacSM{G}{\cdot}$ are constructed similarly as $\mat{X}_k$.
Refer to \cite[Sec. 2]{sylgkf5seg2020} for definition of $\LG{SE(3)}$ and $\LG{\R^n}$ operators.

\subsection{Lie group constrained EKF (LG-CEKF)}
The \textit{a priori}, \textit{a posteriori}, and constrained state mean estimates for time step $k$ are denoted by $\kfsp{\mu}{k}$, $\kfsm{\mu}{k}$, and $\kfsc{\mu}{k}$, respectively. 
    Note that the true state $\mat{X}_k$ can be expressed as $\vec{\mu}_{k} \LvectranSM{G}{\boldsymbol{\epsilon}}$, also denoted as $\vec{\mu}_{k}^{\boldsymbol{\epsilon}}$ (note of superscript $\boldsymbol{\epsilon}$), where $\vec{\mu}_{k}$ is one of the estimated state means just mentioned with error, $\vec{\epsilon}$.
The estimated KF state error \textit{a priori} and \textit{a posteriori} covariance matrices are denoted as $\kfcp{P}{k}$ and $\kfcm{P}{k}$, respectively.
    Note, the error covariance is not updated at the constraint update step.
The KF is based on the Lie group EKF, as defined in \cite{Bourmaud2013}.

\subsubsection{Prediction step} \label{sec:pred-update}
estimates the \textit{a priori} state $\kfsp{\mu}{k}$ at the next time step and may not necessarily respect the kinematic constraints of the body, so joints may become dislocated after this prediction step.
The mean propagation of the three instrumented body segments is governed by Eq. \eqref{eq:lgkf-predmu} 
    where $\Omega (\mat{X}_{k\smallneg1},\vec{u}_{k})$ is the motion model for the tracked body segments,
    and input $\vec{u}_k$ contains acceleration with respect world frame $W$ and angular velocity with respect body frame, 
    as obtained by the IMU attached to segment $b$ (denoted as $\bvmeas{W}{b}{a}{}{k}$ and $\bvmeas{b}{}{\omega}{}{k}$ for $b \in \{p, lf, rf\}$).
For the sake of brevity, only the motion model of the position, orientation, and velocity for body segment $b$ is shown (Eq. \eqref{eq:lgkf-omegab}).
    The complete $\Omega (\kfsc{\mu}{k\smallneg1},\vec{u}_{k})$ contains the motion model for body segments $\{p, lf, rf\}$.
Note that one may integrate the measured angular velocity, $\bvmeas{b}{}{\omega}{}{k}$, to predict orientation.
    However, we set the angular velocity input to zero (\ie{} $\bvmeas{b}{}{\omega}{}{k} = 0$) to simplify computations, knowing that the orientation will be updated in the measurement step using measurements from a third-party orientation estimation algorithm which integrates angular velocity.
\begin{gather}
	\kfsp{\mu}{k} = \kfsc{\mu}{k-1} \LvectranSM{G}{\kfsc{\Omega}{k}}, \text{ where } \kfsc{\Omega}{k} = \Omega (\kfsc{\mu}{k\smallneg1},\vec{u}_{k}) \label{eq:lgkf-predmu} \\
	\Omega^{b} (\kfsc{\mu}{k\smallneg1},\vec{u}_{k}) = \left[\begin{array}{c}
        	\kfbvc{W}{b}{R}{T}{k\smallneg1} (
        	\dt \kfbvc{W}{b}{v}{}{k\smallneg1} +
        	\tfrac{\dt^2}{2} \bvmeas{W}{b}{a}{}{k} 
        	) \\
        	\dt \bvmeas{b}{}{\omega}{}{k} \\
        	\dt \bvmeas{W}{b}{a}{}{k} 
    \end{array} \right] \label{eq:lgkf-omegab}
\end{gather}

The state error covariance matrix propagation is governed by Eq. \eqref{eq:lgkf-predP}, 
    where $\mathcal{F}_{k}$ represents the matrix Lie group equivalent to the Jacobian of $f(\vec{X}_{k\smallneg1},\vec{u}_{k},\vec{n}_{k})$,
    $\mathcal{Q}$ is the covariance matrix of the process noise,
    and $\mathscr{C}_{k}
        = \tfrac{\partial}{\partial \vec{\epsilon}} \Omega( \kfsec{\mu}{k-1}, \vec{u}_{k} ) |_{\vec{\epsilon} = 0}$ represents the linearization of the motion model with an infinitesimal perturbation $\vec{\epsilon}$.
The process noise covariance matrix, $\mathcal{Q}$, is calculated from the input-to-state matrix $\mathcal{G}$ (\ie{} a $27 \times 9$ matrix with values $\tfrac{\Delta t^2}{2} \mat{I}_{3 \times 3}$ for corresponding position states, and values $\Delta t \mat{I}_{3 \times 3}$ for corresponding velocity states) and the noise variances of the measured acceleration and angular velocity, $\bv{}{}{\sigma}{2}{a}$ and $\bv{}{}{\sigma}{2}{\omega}$, respectively.
Refer to the supplementary material \cite{sylgkf7segsupp2020} for the explicit definition of $\Omega (\kfsc{\mu}{k\smallneg1},\vec{u}_{k})$, $\mathcal{G}$, and $\mathscr{C}_{k}$.
\begin{gather}
    \kfcp{P}{k} = \mathcal{F}_{k} \kfcm{P}{k-1} \mathcal{F}_{k}^T + \LJacSM{G}{\kfsc{\Omega}{k}} \mathcal{Q} \LJacSM{G}{\kfsc{\Omega}{k}}^T \label{eq:lgkf-predP} \\
    \mathcal{F}_{k} = \LAdSM{G}{\LvectranSM{G}{\smallneg\kfsc{\Omega}{k}}} + \LJacSM{G}{\kfsc{\Omega}{k}} \mathscr{C}_k \\
    \mathcal{Q} = \mathcal{G} \: \diag (
       	    \bv{}{}{\sigma}{2}{a}, \bv{}{}{\sigma}{2}{\omega}
       	) \: \mathcal{G}^T ,\quad
    \mathscr{C}_{k} = \tfrac{\partial}{\partial \vec{\epsilon}} 
    \Omega( \kfsec{\mu}{k-1}, \vec{u}_{k} ) |_{\vec{\epsilon} = 0} \label{eq:lgkf-omeganoise}  \\
    \LJacSM{G}{\vec{v}} = \textstyle\sum_{i=0}^{\infty} \tfrac{(-1)^i}{(i+1)!} \LadSM{G}{\vec{v}}^i \text{ , } \vec{v} \in \R^n \label{eq:ljacG}
\end{gather}

\subsubsection{Measurement update} \label{sec:meas-update}
estimates the state at the next time step through: 
    (a) orientation update (\emph{ori}), 
    (b) mid-pelvis pseudo-measurements (\emph{mp}), 
    (c) $z$-position assumptions (\emph{mpz}/\emph{lfz}/\emph{rfz}),
    (d) pelvis yaw pseudo-measurement (\emph{yaw}),
	and (e) foot zero-velocity update (\emph{lfv}/\emph{rfv}).
    Each type of measurement will be described later.
The \textit{a posteriori} state $\kfsm{\mu}{k}$ and the covariance $\kfcm{P}{k}$ are calculated following the Lie EKF equations below.
    $\mathcal{H}_{k}$ can be seen as the matrix Lie group equivalent to the Jacobian of $h(\mat{X}_k)$
	    and is defined as the concatenation of $\mathcal{H}_{ori}$, $\mathcal{H}_{mp,k}$, $\mathcal{H}_{mpz,k}$, $\mathcal{H}_{lfz,k}$, and $\mathcal{H}_{rfz,k}$. 
	    $\mathcal{H}_{yaw,k}$ is also concatenated to $\mathcal{H}_{k}$ for \LGKFgSegbIMU{} (\ie{} no pelvis measurement is available).
        $\mathcal{H}_{lfv}$ and/or $\mathcal{H}_{rfv}$ are concatenated to $\mathcal{H}_{k}$ when the left and/or right foot contact is detected.
$\mat{Z}_k$, $h\left( \mat{X}_k \right)$, and $\mathcal{R}_k$ are constructed similarly to $\mathcal{H}_{k}$ but combined using $\diag$ instead of concatenation 
	(\eg{} $\mathcal{R}_{k} = \diag(\bv{}{}{\sigma}{2}{ori}, \bv{}{}{\sigma}{2}{mp})$).
\begin{gather}
    \kfsm{\mu}{k} = \kfsp{\mu}{k} \LvectranSM{G}{\vec{\nu}_k} ,\:\:
    \vec{\nu}_{k} = \mat{K}_{k} \LtranvecSM{G_m}{h(\kfsp{\mu}{k})^{-1} \mat{Z}_k} \\
    \mat{K}_{k} = \kfcp{P}{k} \mathcal{H}_{k}^T ( \mathcal{H}_{k} \kfcp{P}{k} \mathcal{H}_{k}^T + \mathcal{R}_{k} )^{-1} \\
    \mathcal{H}_{k} = \tfrac{\partial}{\partial \vec{\epsilon}} \Ltranvec{G_m}{
    	h( \kfsp{\mu}{k} )^{-1} 
    	h( \kfsep{\mu}{k} ) } |_{\vec{\epsilon}=0} \label{eq:dHdef} \\
    \kfcm{P}{k} = \LJac{G}{\vec{\nu}_k} \left(\mat{I} - \mat{K}_{k} \mathcal{H}_{k} \right) \kfcp{P}{k} \LJac{G}{\vec{\nu}_k}^T \label{eq:Ptildek}
\end{gather}

\paragraph{Orientation update}
    uses the new orientation measurements of body segments from IMUs, denoted as $\bvmeas{W}{p}{R}{}{k}$, $\bvmeas{W}{lf}{R}{}{k}$, and $\bvmeas{W}{rf}{R}{}{k}$.
The measurement function is shown in Eqs. \eqref{eq:h-ori-k}-\eqref{eq:Z-ori-k} with measurement noise variance $\bv{}{}{\sigma}{2}{ori}$ ($9 \times 1$ vector).
$\mat{I}_{i \times i}$ and $\mat{0}_{i \times j}$ denote $i \times i$ identity and $i \times j$ zero matrices.
$\mathcal{H}_{ori}$ (Eq. \eqref{eq:H-ori-k}), as well as any other $\mathcal{H}_{a}$ for some measurement function $a$, are calculated by applying Eq. \eqref{eq:dHdef} to their corresponding measurement function, 
    followed by tedious algebraic manipulation (\eg{} using the property $\Lhat{}{\vec{a}}\vec{b} = \Lodot{}{\vec{b}}\vec{a}$ as defined in \cite[][Eq. (72)]{barfoot2017state})
    and first order linearization (\ie{} $\exp(\Lhat{}{\boldsymbol{\epsilon}}) \approx \mat{I} + \Lhat{}{\boldsymbol{\epsilon}}$).
Note that for \LGKFgSegbIMU{}, pelvis orientation is not updated (\ie{} first row of $\mathcal{H}_{ori}$ is omitted).
\begin{gather}
    h_{ori} \left( \mat{X}_{k} \right) = \diag(\bv{W}{p}{R}{}{k}, \bv{W}{lf}{R}{}{k}, \bv{W}{rf}{R}{}{k}) \label{eq:h-ori-k} \\
	\mat{Z}_{ori} = \diag(\bvmeas{W}{p}{R}{}{k}, \bvmeas{W}{lf}{R}{}{k}, \bvmeas{W}{rf}{R}{}{k}) \label{eq:Z-ori-k} \\[12pt]
	\mathcal{H}_{ori} = \left[ \begin{array}{ccc:c}
	    \bovermat{segment poses}{
		\mat{0}_{3 \times 3} \: \mat{I}_{3 \times 3} & \phantom{\mat{0}_{3 \times 3} \: \mat{I}_{3 \times 3}} & \phantom{\mat{0}_{3 \times 3} \: \mat{I}_{3 \times 3}} } & \bovermat{vel.}{\phantom{\mat{0}_{9 \times 9}}} \\
		& \mat{0}_{3 \times 3} \: \mat{I}_{3 \times 3} & & \mat{0}_{9 \times 9} \\
    	& & \mat{0}_{3 \times 3} \: \mat{I}_{3 \times 3} & \\
    	\end{array} \right] \label{eq:H-ori-k}
\end{gather}
    
\paragraph{Mid-pelvis pseudo-measurement}
    enforces a mid-pelvis pseudo-measurement where the pelvis $x$ and $y$ position is the approximate average of the $x$ and $y$ position of the two feet.
The measurement function is shown in Eqs. \eqref{eq:Z-mp-k}-\eqref{eq:h-mp-k}, with measurement noise variance $\bv{}{}{\sigma}{2}{mp}$ ($2 \times 1$ vector).
    $\vec{i}_x$, $\vec{i}_y$, $\vec{i}_z$, and $\vec{i}_0$ denote $4 \times 1$ selector vectors whose 1$^{st}$ to 4$^{th}$ row, respectively, are $1$, while the rest are $0$ (\eg{} $\bv{W}{ls}{T}{}{k} \vec{i}_{z}$ returns the long axis of the left shanks).
$\mathcal{H}_{mp}$ (Eq. \eqref{eq:H-mp-k}) is derived similarly as \cite[Eq. (36)]{sylgkf5seg2020}.
\begin{gather}
    \mat{E}_{2} = \begin{bmatrix}
	    \mat{I}_{2 \times 2} & \mat{0}_{2 \times 2}
	\end{bmatrix}, \quad 
	\Ltranvec{}{\mat{Z}_{mp}} = \mat{0}_{2 \times 1} \label{eq:Z-mp-k}\\
	\LtranvecKFFunc{}{h}{}{mp}{\mat{X}_{k}} = \mat{E}_{2} ( 
	    \bv{W}{p}{T}{}{}
	    - \tfrac{1}{2} \bv{W}{lf}{T}{}{k}
	    - \tfrac{1}{2} \bv{W}{rf}{T}{}{k} ) \bv{}{}{i}{}{0} \label{eq:h-mp-k} \\
\mathcal{H}_{mp,k} =  \left[ \begin{array}{c:c:c:c}
    	\mat{E}_{2} \kfbvp{W}{p}{T}{}{k} \Lodot{}{\vec{i}_0} &
    	\smallneg \tfrac{ \mat{E}_{2} \kfbvp{W}{lf}{T}{}{k} \Lodot{}{\vec{i}_0} }{2}  &
    	\smallneg \tfrac{ \mat{E}_{2} \kfbvp{W}{rf}{T}{}{k} \Lodot{}{\vec{i}_0} }{2}  &
    	\mat{0}_{2 \times 9}
    \end{array} \right] \label{eq:H-mp-k}
\end{gather}
\paragraph{$z$-position assumptions}
	bring the pelvis $z$ position to initial pelvis height, $z_{p}$,
	and the foot $z$ position strictly close to the floor level, $z_{f}$, when a foot step is detected, but is gradually relaxed as time passes (\eg{} relaxed after 1 second after the foot step).
The measurement function for the left foot is shown in Eq. \eqref{eq:hZ-lfz-k}, with measurement noise variance $\bs{}{}{\sigma}{2}{lfz,k}$ (varies with time and decays to $0.02$ of initial value, $\bs{}{}{\sigma}{}{lfz0}$, after $\lambda$ seconds as shown in Eq. \eqref{eq:sigma-lfz-k}).
$k_{s}$ is the time step of the last foot step detected.
$\mathcal{H}_{lfz}$ (Eq. \eqref{eq:H-lfz-k}) is derived similarly as \cite[Eq. (38)]{sylgkf5seg2020}.
    The right foot and mid-pelvis $z$-position assumption can be calculated similarly,
    except that for the mid-pelvis $z$-position, the measurement noise variance, $\bv{}{}{\sigma}{2}{mp}$, is constant, and $\Ltranvec{}{\mat{Z}_{mpz}} = z_p$.
\begin{gather}
    \LtranvecKFFunc{}{h}{}{lfz}{\mat{X}_{k}} = \vec{i}_z^T \bv{W}{lf}{T}{}{k} \vec{i}_0, \quad
	\Ltranvec{}{\mat{Z}_{lfz}} = z_f \label{eq:hZ-lfz-k} \\
	\bs{}{}{\sigma}{}{lfz,k} = \bs{}{}{\sigma}{}{lfz0} \exp{(\nicefrac{-4(k-k_{s})\dt}{\lambda})} \label{eq:sigma-lfz-k} \\
    \mathcal{H}_{lfz,k} = \left[ \begin{array}{c:c:c:c}
    		\mat{0}_{1 \times 6} & 
    		\vec{i}_z^T \kfbvp{W}{lf}{T}{}{k} \Lodot{}{\vec{i}_0} & 
    		\mat{0}_{1 \times 6} & 
    		\mat{0}_{1 \times 9} \\
    	\end{array} \right] \label{eq:H-lfz-k}
\end{gather}

\paragraph{Pelvis yaw pseudo-measurement}
    encourages the pelvis yaw orientation to be the average of the yaw orientations of both feet. 
    Pelvis pitch and roll pseudo-measurements are set to zero.
This pseudo-measurement is only used in \LGKFgSegbIMU{} (\ie{} when the pelvis IMU measurement is not available).
The measurement function (Eqs. \eqref{eq:h-yaw-k}-\eqref{eq:Z-yaw-k}) denotes that pelvis orientation equals the rotation along unit vector $\vec{\alpha}$ (\ie{} $z$ axis) by $\theta_{yaw}(\bv{W}{lf}{R}{}{k},\bv{W}{rf}{R}{}{k})$ radians,
    with measurement noise variance $\bv{}{}{\sigma}{2}{yaw}$ ($3 \times 1$ vector).
    $\theta_{yaw}(\bv{W}{lf}{R}{}{k},\bv{W}{rf}{R}{}{k})$ (Eq. \eqref{eq:theta-yaw-k}) was calculated from the inverse tangent of the resultant vector of the $z$ axes of both feet (\ie{} $(\bv{W}{lf}{R}{}{k}+\bv{W}{rf}{R}{}{k}) \bv{}{}{i}{}{3z}$).
    Note that the resultant vector also divides the angle between the feet $z$ axes (\ie{} long axes) equally (\ie{} bisector).
$\bv{}{}{i}{}{3x}$, $\bv{}{}{i}{}{3y}$, and $\bv{}{}{i}{}{3z}$ denote $3 \times 1$ selector vectors whose 1$^{st}$ to 3$^{rd}$ row, respectively, are $1$, while the rest are $0$.
    Refer to the supplementary material \cite{sylgkf7segsupp2020} for the derivation of $\mathcal{H}_{yaw,k}$.
\begin{gather}
    \theta_{yaw}(\bv{W}{lf}{R}{}{k},\bv{W}{rf}{R}{}{k}) = \tan^{-1} \Big( 
        \tfrac{ \bv{}{}{i}{T}{3y} (\bv{W}{lf}{R}{}{k}+\bv{W}{rf}{R}{}{k}) \bv{}{}{i}{}{3z} }
             { \bv{}{}{i}{T}{3x} (\bv{W}{lf}{R}{}{k}+\bv{W}{rf}{R}{}{k}) \bv{}{}{i}{}{3z} }  \Big) 
        \in \R \label{eq:theta-yaw-k} \\
    \LKFFunc{h}{}{yaw}{\mat{X}_{k}} = \LvectranSM{}{ \overbrace{\theta_{yaw}(\bv{W}{lf}{R}{}{k},\bv{W}{rf}{R}{}{k}) \vec{\alpha}}^{\approx\text{ mean yaw of two feet}} }^{T} \bv{W}{p}{R}{}{k} \label{eq:h-yaw-k} \\ 
    \vec{\alpha} = \begin{bmatrix} 0 & 0 & 1 \end{bmatrix}^T ,\quad
    \mat{Z}_{yaw} = \mat{I}_{3 \times 3} \label{eq:Z-yaw-k} \\
    \mathcal{H}_{yaw,k} = \tfrac{\partial}{\partial \bv{}{}{\epsilon}{}{}} \LtranvecSM{}{
			\LKFFunc{h}{}{yaw}{\kfsp{\mu}{k}}^{-1} 
			h_{yaw} ( \kfsep{\mu}{k} ) } |_{\bv{}{}{\epsilon}{}{} = 0}
\end{gather}

\paragraph{Foot zero-velocity update (ZUPT)}
    encourages foot velocity to approach zero when a foot step is detected.
The measurement function for the ZUPT at the left foot is shown in Eq. \eqref{eq:hZ-lfv-k} with measurement noise variance $\bv{}{}{\sigma}{2}{lfv}$ ($3 \times 1$ vector).
    The right foot ZUPT can be calculated similarly.
\begin{gather}
    \LtranvecKFFunc{}{h}{}{lfv}{\mat{X}_{k}} = \bv{W}{lf}{v}{}{}, \quad
	\Ltranvec{}{\mat{Z}_{lfv}} = \mat{0}_{3 \times 1} \label{eq:hZ-lfv-k} \\
    \mathcal{H}_{lfv} = \left[ \begin{array}{cccc}
		\mat{0}_{3 \times 18} & 
		\mat{0}_{3 \times 3} &
		\mat{I}_{3 \times 3} & 
		\mat{0}_{3 \times 3} \\
	\end{array} \right] \label{eq:H-lfv-k}
\end{gather}

\subsubsection{Satisfying biomechanical constraints} \label{sec:const-nonlin-update}
After the prediction and measurement updates, above, the body joints may have become dislocated.
    This update corrects the kinematic state estimates to satisfy the biomechanical constraints of the human body by projecting the current \textit{a posteriori} state $\kfsm{\mu}{k}$ estimate onto the constraint surface, guided by our uncertainty in each state variable, encoded by $\kfcm{P}{k}$.
The constraint step enforces the following biomechanical limitations:
    (a) hinged knee and ankle joints (\emph{lj}/\emph{rj}),
    and (b) maximum leg length (\emph{ll}/\emph{rl}).
The constrained state, $\kfsc{\mu}{k}$, can be calculated using the equations below, similar to the measurement update of \cite{Bourmaud2013} but with zero noise,
    and $\mathcal{C}_{k} = \begin{bmatrix} \mathcal{C}_{L, k}^T & \mathcal{C}_{R, k}^T \end{bmatrix}^T$.
$\mathcal{C}_{L, k}$ is the concatenation of $\mathcal{C}_{lj,k}$ and $\mathcal{C}_{ll,k}$; 
	the latter matrix implements an inequality constraint and is only concatenated when the distance between the ankle and hip is greater than the leg length, $d^{lt}+d^{ls}$, after the preceding measurement update step.
$\mathcal{C}_{R, k}$ can be derived similarly,
	while $\mat{D}_k$ and $c \left( \mat{X}_k \right)$ are constructed similarly to $\mat{Z}_k$.
\begin{gather}
    \kfsc{\mu}{k} = \kfsm{\mu}{k} \exp_{G} \left( \Lhat{G}{\vec{\nu}_k} \right) \\
    \vec{\nu}_{k} = \mat{K}_{k} ( \LveeSM{G_c}{\log_{G_c} ( c(\kfsm{\mu}{k})^{-1} \mat{D}_k ) }) \\
    \mat{K}_{k} = \kfcm{P}{k} \mathcal{C}_{k}^T ( \mathcal{C}_{k} \kfcm{P}{k} \mathcal{C}_{k}^T )^{-1} \\
    \mathcal{C}_{k} = \tfrac{\partial}{\partial \vec{\epsilon}} \LveeSM{G_c}{
    	\log_{G_c} \big( 
    	c\left( \kfsm{\mu}{k} \right)^{-1} 
    	c\left( \kfsem{\mu}{k} \right) \big) 
    	} |_{\vec{\epsilon}=0} \label{eq:dCdef}
\end{gather}
    
\paragraph{Hinged knee and ankle joints constraints}
    forces the knees and ankles to act as hinge joints, as defined in Eqs. \eqref{eq:c-lleghinge}-\eqref{eq:D-lleghinge},
    where $\bv{W}{l}{\tau}{}{}(\kfsc{\mu}{k})$ (Eq. \eqref{eq:thigh-vect}) denotes the left ankle-to-hip vector \changed{whose dot product with the foot $y$ axis,} $\bv{W}{lf}{r}{}{y} = \mat{E} \bv{W}{lf}{T}{}{k} \bv{}{}{i}{}{y}$, \changed{equals zero}.
$\mathcal{C}_{lj,k}$ (Eq. \eqref{eq:C-lleghinge}) is derived similarly as \cite[Eq. (66)]{sylgkf5seg2020} and is defined explicitly in the supplementary material \cite{sylgkf7segsupp2020}.
    Note that the sensor attached to the feet is assumed to be located between the toe and the heel.
    Lastly, the right side (\ie{} $\mathcal{C}_{rj,k}$) can be derived similarly.
\begin{gather}
    \bv{p}{lh}{p}{}{} = \begin{bmatrix}
        0 & \nicefrac{d^{\cs{p}}}{2} & 0 & 1
        \end{bmatrix}^T ,\quad
    \bv{lf}{la}{p}{}{} = \begin{bmatrix}
        0 & 0 & \nicefrac{d^{\cs{lf}}}{2} & 1
        \end{bmatrix}^T \\
    \bv{W}{l}{\tau}{}{}(\mat{X}_{k}) = 
    	\overbrace{\begin{bmatrix}
    		\mat{I}_{3 \times 3} & \mat{0}_{3 \times 1}
    		\end{bmatrix}}^{\mat{E}} \big(
        \overbrace{\bv{W}{p}{T}{}{k} \: \bv{p}{lh}{p}{}{} }^{\text{hip joint pos.}} - 
        \overbrace{\bv{W}{lf}{T}{}{k} \: \bv{lf}{la}{p}{}{} }^{\text{ankle joint pos.}} \big) \label{eq:thigh-vect} \\
    \LtranvecKFFunc{}{c}{}{lj}{\mat{X}_{k}} = (\mat{E}\bv{W}{lf}{T}{}{k}\bv{}{}{i}{}{y})^{T} \bv{W}{l}{\tau}{}{}(\mat{X}_{k}) \label{eq:c-lleghinge} \\
	\Ltranvec{}{\mat{D}_{lj}} = 0  \label{eq:D-lleghinge} \\
	\mathcal{C}_{lj,k} = \tfrac{\partial}{\partial \bv{}{}{\epsilon}{}{}} \LtranvecSM{}{
			\LKFFunc{c}{}{lj}{\kfsm{\mu}{k}}^{-1} 
			\LKFFunc{c}{}{lj}{\kfsem{\mu}{k}} } |_{\bv{}{}{\epsilon}{}{} = 0} \label{eq:C-lleghinge}
\end{gather}

\paragraph{Leg length constraint}
    enforces that the distance between the ankles and hips ($||\bv{W}{l}{\tau}{}{}(\mat{X}_{k})||$ and $||\bv{W}{r}{\tau}{}{}(\mat{X}_{k})||$) cannot be more than the leg length, $d^{\cs{lt}}+d^{\cs{ls}}$, as defined in Eqs. \eqref{eq:c-lleglength}-\eqref{eq:D-lleglength}.
$\mathcal{C}_{ll,k}$ (Eq. \eqref{eq:C-lleglength}) is derived similarly as \cite[Eq. (44)]{sylgkf5seg2020} and is defined explicitly in the supplementary material \cite{sylgkf7segsupp2020}.
    The right side (\ie{} $\mathcal{C}_{rl,k}$) can be derived similarly.
\begin{gather}
    \LtranvecKFFunc{}{c}{}{ll}{\mat{X}_{k}}
    	= \left(\bv{W}{l}{\tau}{}{}(\mat{X}_{k})\right)^T 			
    \bv{W}{l}{\tau}{}{}(\mat{X}_{k}) \label{eq:c-lleglength} \\
    \Ltranvec{}{\mat{D}_{ll}} = (d^{\cs{lt}}+d^{\cs{ls}})^2 \label{eq:D-lleglength} \\
    \mathcal{C}_{ll,k} = \tfrac{\partial}{\partial \bv{}{}{\epsilon}{}{}} \LtranvecSM{}{
			\LKFFunc{c}{}{ll}{\kfsm{\mu}{k}}^{-1} 
			\LKFFunc{c}{}{ll}{\kfsem{\mu}{k}} } |_{\bv{}{}{\epsilon}{}{} = 0} \label{eq:C-lleglength}
\end{gather}

\subsection{Post-processing}
The thigh and shank orientations were estimated under the assumption that both knees and ankles are hinge joints (\ie{} $\bv{W}{lf}{r}{}{y}$ = $\bv{W}{ls}{r}{}{y}$ = $\bv{W}{lt}{r}{}{y}$).
    Refer to Fig. \ref{fig:body-skeleton} for visualization.
The angle between the left shank segment and hip-to-ankle vector $\bv{W}{l}{\tau}{}{}(\kfsc{\mu}{k})$, $\bs{}{l}{\theta}{}{k}$, can be solved using the cosine law, as shown in Eq. \eqref{eq:theta-lah-coslaw}.
    The left shank normal axis, $\kfbvc{W}{ls}{r}{}{z,k}$, is then estimated by rotating $\bv{W}{l}{\tau}{}{}(\kfsc{\mu}{k})$ through $\kfbvc{W}{lf}{r}{}{y,k}$ by $\bs{}{l}{\theta}{}{k}$ degrees, as shown in Eq. \eqref{eq:theta-lah-lsrz}.
    Finally, the left shank and thigh orientation are calculated using Eqs. \eqref{eq:postproc-plk-wltrz}-\eqref{eq:postproc-WlstR}, respectively.
The right side is calculated similarly.
Note that $\LhatSM{}{\vec{r}_1} \vec{r}_2$ is equal to the cross product of $\vec{r}_1$ and $\vec{r}_2$.

\begin{gather}
    \bs{W}{lt}{d}{2}{} = \bs{W}{ls}{d}{2}{} + ||\bv{W}{l}{\tau}{}{}(\kfsc{\mu}{k})||^2 - 2 \bs{W}{ls}{d}{}{} ||\bv{W}{l}{\tau}{}{}(\kfsc{\mu}{k})|| \cos{(\bs{}{l}{\theta}{}{k})}  \nonumber \\
    \bs{}{l}{\theta}{}{k} = \cos^{-1} \left( \tfrac{\bs{W}{ls}{d}{2}{} + ||\bv{W}{l}{\tau}{}{}(\kfsc{\mu}{k})||^2 - \bs{W}{lt}{d}{2}{}}{2 \bs{W}{ls}{d}{}{} ||\bv{W}{l}{\tau}{}{}(\kfsc{\mu}{k})||} \right) \label{eq:theta-lah-coslaw} \\
    \kfbvc{W}{ls}{r}{}{z,k} = \bv{}{l}{R}{}{k} \: \tfrac{\bv{W}{l}{\tau}{}{}(\kfsc{\mu}{k})}{||\bv{W}{l}{\tau}{}{}(\kfsc{\mu}{k})||} 
    \text{ where } \bv{}{l}{R}{}{k} = \LvectranSM{}{ \bs{}{l}{\theta}{}{k} \kfbvc{W}{lf}{r}{}{y,k}} \label{eq:theta-lah-lsrz} \\
    \kfbvc{W}{lt}{r}{}{z,k} = 
        \tfrac{\kfbvc{W}{lh}{p}{}{k} - \kfbvc{W}{lk}{p}{}{k}}
        {||\kfbvc{W}{lh}{p}{}{k} - \kfbvc{W}{lk}{p}{}{k}||} \text{ where}\: \kfbvc{W}{lk}{p}{}{k} = \kfbvc{W}{lf}{T}{}{k} \bv{lf}{la}{p}{}{} \small{+}
        d^{ls} \kfbvc{W}{ls}{r}{}{z,k} 
        \label{eq:postproc-plk-wltrz} \\
    \kfbvc{W}{ls}{R}{}{k} = \left[ \begin{array}{c:c:c}
        \LhatSM{}{\kfbvc{W}{lf}{r}{}{y,k}} \kfbvc{W}{ls}{r}{}{z,k} &
        \kfbvc{W}{lf}{r}{}{y,k} &
        \kfbvc{W}{ls}{r}{}{z,k}
        \end{array} \right] \\
    \kfbvc{W}{lt}{R}{}{k} = \left[ \begin{array}{c:c:c}
        \LhatSM{}{\kfbvc{W}{lf}{r}{}{y,k}} \kfbvc{W}{lt}{r}{}{z,k} & 
        \kfbvc{W}{lf}{r}{}{y,k} &
        \kfbvc{W}{lt}{r}{}{z,k}
        \end{array} \right] \label{eq:postproc-WlstR}
\end{gather}

\section{Experiment}
\label{sec:experiment-intro}
The algorithm \LGKFgSeg{} was evaluated on two data sets, NeuRA (NR) \cite{syckf2020} and Total Capture Dataset (TCD) \cite{Trumble2017}, as described in Table \ref{tab:db-benchmark-conf}.
    Raw data were captured using a commercial IMC (\ie{} Xsens Awinda) compared against a benchmark OMC (\ie{} Vicon).
    Two benchmark models were generated from the NeuRA data set: 
        i) the conventional gait model generated from Vicon's Plug-in Gait (PiG);
        ii) and a kinematically-fitted model (KFit) from the Vicon Nexus pipeline.
    A notable difference between the PiG and KFit model is that the KFit model inherently assumes a 1 DoF hinged knee joint and constant segment length when reconstructing kinematics.
    The benchmark model from the TCD data set was obtained from Vicon Blade and is similar to the KFit model.  
The algorithm was evaluated on movements listed in Table \ref{tab:movement-type-desc}.
    
\begin{table}[htbp]
  \centering
  \caption{Dataset and Benchmark configurations}
    \begin{tabular}{p{3em}p{3em}ccp{5.145em}ccc}
    \hline
    \hline
    \multirow{2}[2]{*}{DB} & \multirow{2}[2]{*}{Model} & \multicolumn{2}{p{3em}}{Subject} & \multirow{2}[2]{5.145em}{Benchmark System} & \multirow{2}[2]{6em}{Capture Area (m$^2$)} & \multirow{2}[2]{*}{IMU} & \multirow{2}[2]{4em}{Sample Rate (Hz)} \bigstrut[t]\\
    \multicolumn{1}{c}{} & \multicolumn{1}{c}{} & \multicolumn{1}{p{1.5em}}{M} & \multicolumn{1}{p{1.5em}}{F} & \multicolumn{1}{c}{} & \multicolumn{1}{c}{} & \multicolumn{1}{c}{} & \bigstrut[b]\\
    \hline
    \hline
    NeuRA & PiG   & \multirow{2}[2]{*}{7} & \multirow{2}[2]{*}{2} & \multirow{2}[2]{*}{Vicon Nexus} & \multirow{2}[2]{*}{\texttildelow$4 \times 4$} & \multirow{2}[2]{3em}{Xsens Awinda} & \multirow{2}[2]{*}{100} \bigstrut[t]\\
    (NR) & KFit  &       &       & \multicolumn{1}{l}{} & \multicolumn{1}{l}{} & \multicolumn{1}{c}{} & \bigstrut[b]\\
    \hline
    TCD   & KFit  & 4     & 1     & Vicon Blade & \texttildelow$4 \times 6$ & Xsens & 60 \bigstrut\\
    \hline
    \hline
    \end{tabular}%
  \label{tab:db-benchmark-conf}%
\end{table}%
\begin{table}[htbp]
    \centering
    \caption{Types of movements done in the validation experiment}
    \label{tab:movement-type-desc}
    \begin{tabular}{clllc}
        \hline \hline
        DB & Movement & Description & Duration & Group \\ \hline 
        \parbox[t]{2mm}{\multirow{7}{*}{\rotatebox[origin=c]{90}{NeuRA}}} & Walk & Walk straight and return & $\sim 30$ s & F\\
         & Figure-of-eight & Walk along figure-of-eight path & $\sim 60$ s & F \\
         & Zig-zag & Walk along zig-zag path & $\sim 60$ s & F \\
         & 5-minute walk & Unscripted walk and stand & $\sim 300$ s & F \\
         & Jog & Jog straight and return & $\sim 30$ s & D \\
         & Jumping jacks & Jumping jacks on the spot & $\sim 30$ s & D \\
         & High-knee jog & High-knee jog on the spot & $\sim 30$ s & D \\ \hline
        \parbox[t]{2mm}{\multirow{3}{*}{\rotatebox[origin=c]{90}{TCD}}} & Walking & Unscripted walk & $\sim 60$ s & F \\
         & Acting & Unscripted walk and acting/standing & $\sim 60$ s & F \\
         & Freestyle & Any movements (\eg{} yoga, crawling) & $\sim 60$ s & D \\
         \hline \hline
    \end{tabular}
    
    F denotes free walk, D denotes dynamic
\end{table}

\subsection{Configuration}
Two variants of \LGKFgSeg{} were tested.
The first variant, \emph{L7S-3I}, takes in input from three IMUs at the pelvis and feet.
The second variant, \emph{L7S-2I}, takes in input from only the two IMUs at the feet,
    where the pelvis input acceleration, $\bvmeas{p}{}{a}{}{k}$, was set as the mean of left and right foot acceleration,
    and the pelvis input angular velocity, $\bvmeas{p}{}{\omega}{}{k}$, was set to zero.

Unless stated, calibration and system parameters similar to \cite{syckf2020, sylgcekf2020} were assumed.
   	The algorithm and calculations were implemented using Matlab 2020a.
    The initial position, orientation, and velocity ($\kfsc{\mu}{0}$) were obtained from the Vicon benchmark system.
    $\kfcm{P}{0}$ was set to $0.5 \mat{I}_{27 \times 27}$.
        The variance parameters used to generate the process and measurement error covariance matrix $\mathcal{Q}$ and $\mathcal{R}$ are shown in Table \ref{tab:var-param-kfnoise}.

    \begin{table}[htbp]
        \centering
        \caption{Variance parameters for generating the process and measurement error covariance matrices, $\mathcal{Q}$ and $\mathcal{R}$ }
        \begin{tabular}{cc|ccccccc} \hline \hline
            \multicolumn{2}{c|}{$\mathcal{Q}$ Parameters} &
            \multicolumn{7}{c}{$\mathcal{R}$ Parameters}
            \\ \hline
            $\bv{}{}{\sigma}{2}{a}$ & 
            $\bv{}{}{\sigma}{2}{\omega}$ &
            $\bv{}{}{\sigma}{2}{ori}$ &
            $\bv{}{}{\sigma}{2}{mp}$ &
            $\bv{}{}{\sigma}{2}{mpz}$ &
            $\bv{}{}{\sigma}{2}{l/rfz0}$ &
            $\lambda$ & 
            $\bv{}{}{\sigma}{2}{yaw}$ & 
            $\bv{}{}{\sigma}{2}{l/rfv}$ \\
            (m$^2$.s$^{-4}$) & 
            (rad$^2$.s$^{-2}$) &
            (rad$^2$) &
            (m$^2$) &
            (m$^2$) &
            (m$^2)$  &
            (s) & 
            (rad$^2$) &
            (m$^2$.s$^{-2}$)
            \\ \hline
            $10^2\mat{1}_{9}$ & 
            $10^7\mat{1}_{9}$ &
            $10\mat{1}_{9}$ &
            $1$ & 
            $1$ & 
            $10$ &
            $1$ & 
            $0.1\mat{1}_{3}$ & 
            $10^{-2} \mat{1}_{3}$ \\
            \hline \hline
        \end{tabular} 
        
        where $\mat{1}_{n}$ is an $1 \times n$ row vector with all elements equal to $1$.  
        \label{tab:var-param-kfnoise}
    \end{table}

\subsection{Metrics}
The evaluation was done using the following metrics.
    Refer to \cite[Sec. III]{syckf2020} for more details.
    
\subsubsection{Mean position and orientation root-mean-square error (RMSE)} 
($e_{pos}$ and $e_{ori}$) both common metrics in video-based human motion capture systems (\eg{} \cite{Marcard2017}). 
In this paper, the set of joint positions $\mathbb{DP}$ is $\{ lh, rh, lk, rk, la, ra, le, re\}$; 
    while the set of uninstrumented body segments $\mathbb{DO}$ is $\mathbb{DO}_{3I} = \{ lt, rt, ls, rs \}$ for \LGKFgSegcIMU{}, and 
    $\mathbb{DO}_{3I} \cup \{ p \}$ for \LGKFgSegbIMU{}.
$\bv{W}{i}{p}{}{k}$ and $\bv{W}{i}{R}{}{k}$ denote the position and orientation of body segment $i$ obtained from the benchmark system.
Note that as the global position of the estimate is still prone to drift due to the absence of an external global position reference, the root position of our system was set equal to that of the benchmark system (\ie{} the  mid-pelvis is placed at the origin in the world frame for all RMSE calculations).
    \begin{gather}
        e_{pos, k} = \tfrac{1}{N_{pos}} \textstyle \sum_{i \in \mathbb{DP}} ||\bv{W}{i}{p}{}{k} - \kfbvc{W}{i}{p}{}{k} || \label{eq:d-pos} \\
        e_{ori, k} = \tfrac{1}{N_{ori}} \textstyle \sum_{i \in \mathbb{DO}} || \LtranvecSM{}{\bv{W}{i}{R}{}{k} \kfbvc{W}{i}{R}{T}{k}} || \label{eq:d-ori}
    \end{gather}

\subsubsection{Hip and knee joint angles RMSE and correlation coefficient (CC)}
The joint angle RMSEs with bias removed (\ie{} the mean difference between the angles over each entire trial was subtracted) and correlation coefficient (CC) of the hip in the sagittal (Y), frontal (X), and transverse (Z) planes, and of the knee in the sagittal (Y) plane.
Note that these joint angles are commonly used parameters in gait analysis.
    
\subsubsection{Spatiotemporal gait parameters}
Specifically, the total travelled distance (TTD), average stride length, and gait speed of the foot are calculated.

\section{Results}
\subsection{Mean position and orientation RMSE}
\begin{figure}[htbp]
        \centering
        \includegraphics[width=\linewidth]{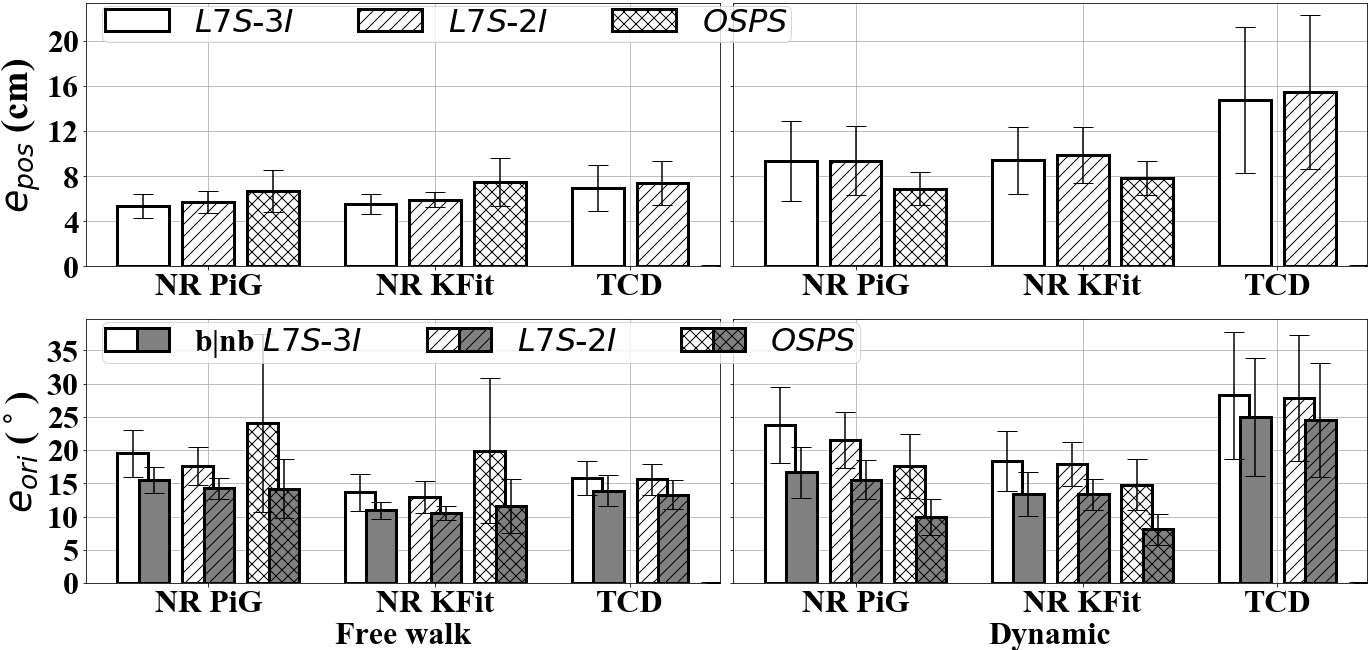}
        \caption{The mean position and orientation RMSE, $e_{pos}$ (top) and $e_{ori}$ (bottom), for \LGKFgSegcIMU{}, \LGKFgSegbIMU{}, and \emph{OSPS} on the NeuRA (NR) PiG and KFit, and TCD database. The prefix \textbf{b} denotes biased, while \textbf{nb} denotes no bias.}
        \label{fig:results-dposdori}
    \end{figure}

    Fig. \ref{fig:results-dposdori} shows the mean position and orientation RMSE of our algorithm for free walking and dynamic movements.
    The comparison involved the output from three algorithms of interest tested on three database configurations (defined in Sec. \ref{sec:experiment-intro}): 
        i-ii) our algorithm using three and two Xsens MTx IMU measurements, respectively, (denoted as \LGKFgSegcIMU{} and \LGKFgSegbIMU{});
        iii) the black box output (\ie{} segment orientation and pelvis position) from the Xsens MVN Studio software (denoted as \emph{OSPS}).
            The \emph{OSPS} result illustrates the performance of a widely-accepted commercial wearable HMCS with an OSPS configuration.
            Note that \emph{OSPS} results are not available for the \emph{TCD} dataset, as neither the result nor the raw files that Xsens MVN requires were present in this dataset.
   
    Both biased and unbiased (\ie{} for unbiased, the mean difference between the angles over each entire trial was subtracted) $e_{ori}$ are presented to account for possible anatomical calibration offset errors between the OMC and OSPS systems \cite{Cloete2008, VanDenNoort2013}.
    
\subsection{Hip and knee joint angle RMSE and CC}
\begin{figure}[htbp]
        \centering
        \includegraphics[width=\linewidth]{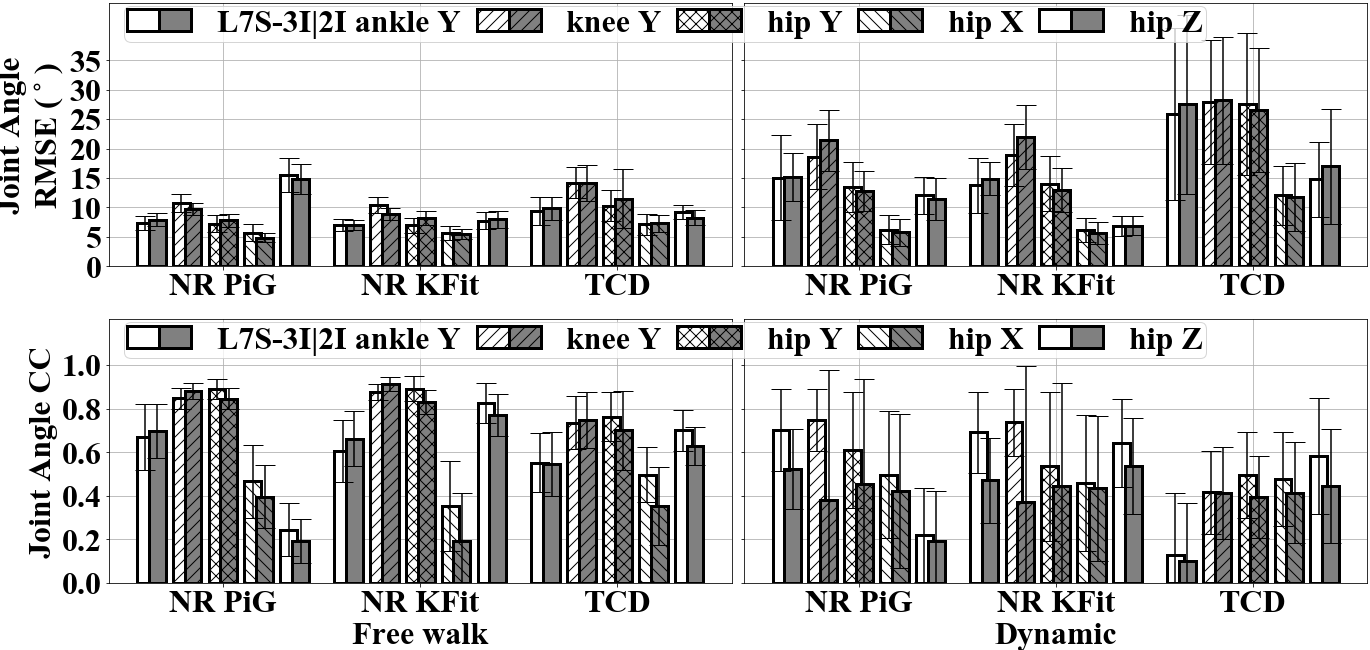}
        \caption{The joint angle RMSE no bias (top) and CC (bottom) of the leg joint angles for \LGKFgSegcIMU{} and \LGKFgSegbIMU{} at each motion type. Y, X, and Z denotes the sagittal, frontal, and transverse plane, respectively.}
        \label{fig:results-kneehip-angles-rmsecc}
    \end{figure}

\begin{figure}[htbp]
        \centering
        \subfloat[Joint angles]{
            \centering
            \includegraphics[width=\linewidth]{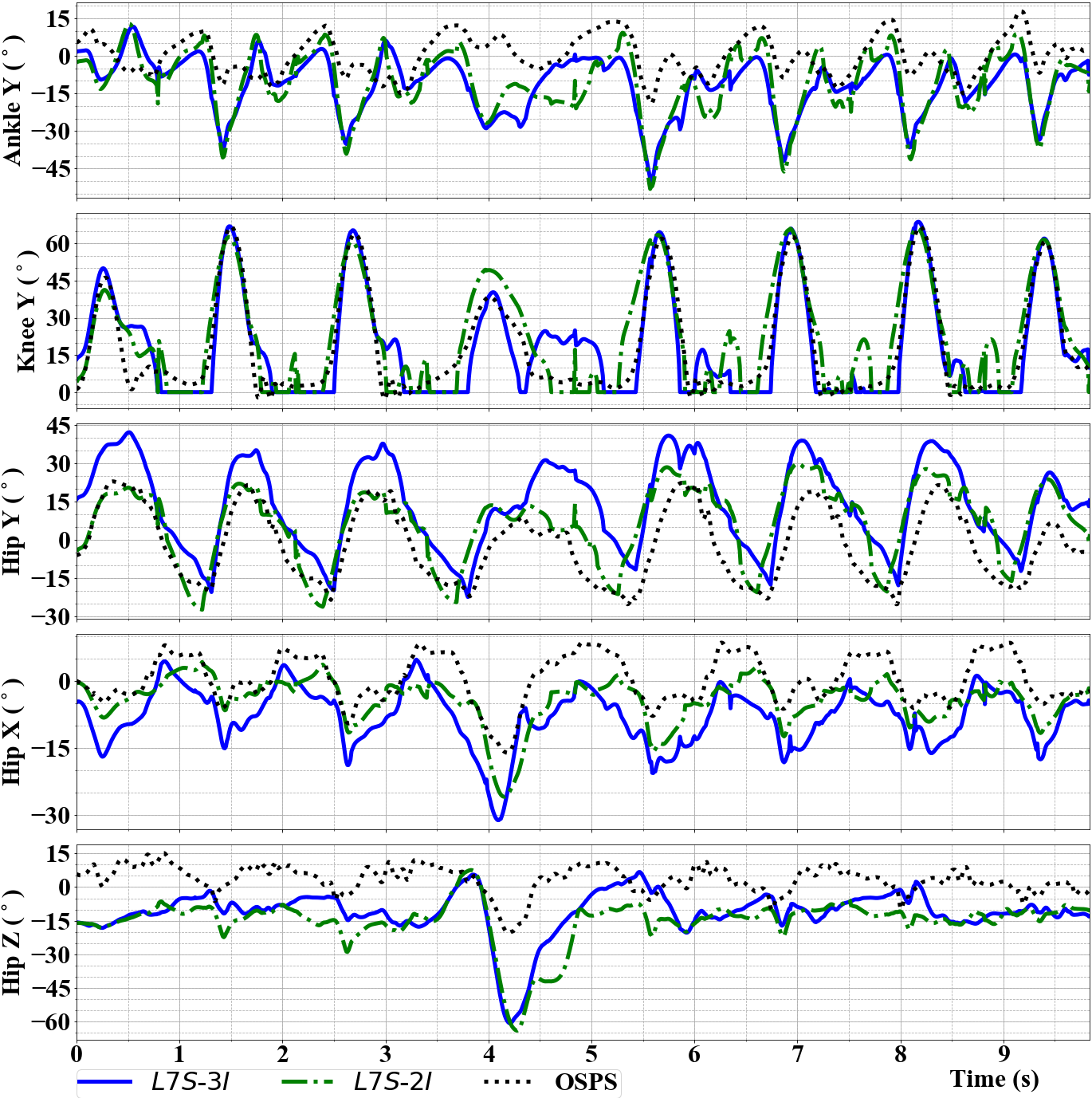}
            \label{fig:results-leg-angle-sample}
        } \\
        \subfloat[$e_{pos}$ and $e_{ori}$]{
            \centering
            \includegraphics[width=\linewidth]{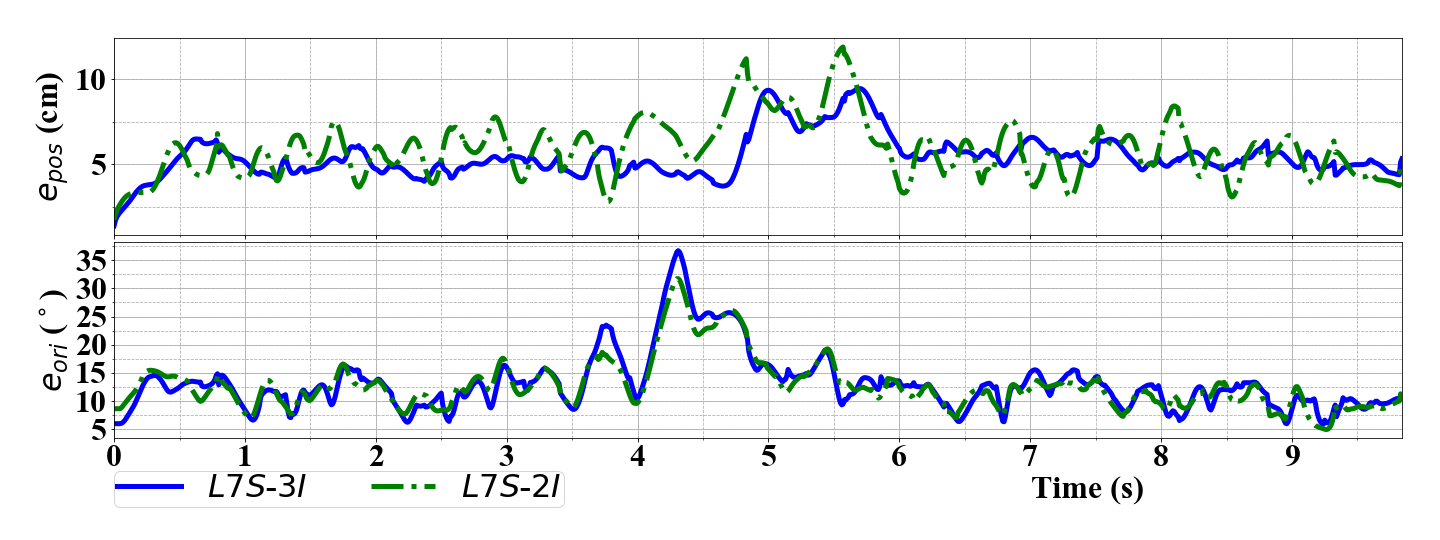}
            \label{fig:results-leg-dposdori}
        }
        \caption{Ankle, knee, and hip joint angle output of \LGKFgSegcIMU{}, \LGKFgSegbIMU{}, and \emph{OSPS} in comparison with the benchmark system (OMC) for a Walk trial. The subject walked straight from $t=0$ to $3$ s, turned $180^\circ$ around from $t=3$ to $5.5$ s, and walked straight to original point from $5.5$ s until the end of the trial.}
        \label{fig:results-kneehip-angle}
    \end{figure}

    Fig. \ref{fig:results-kneehip-angles-rmsecc} shows the ankle, knee, and hip joint angle RMSE (no bias) and CC for \LGKFgSegcIMU{}, \LGKFgSegbIMU{} and \emph{OSPS}.
        Y, X, and Z refers to the plane defined by the normal vectors $y$, $x$, and $z$ axes, respectively,
            and are also known as the sagittal, frontal, and transverse plane in the context of gait analysis.
    Fig. \ref{fig:results-kneehip-angle} shows a sample Walk trial from the NR PiG database.
    \vspace{-6pt}

\subsection{Spatiotemporal gait parameters}

    Table \ref{tab:results-ttddev-sl-gs} shows the TTD, stride length, and gait speed accuracy computed from the global foot position estimate of \LGKFgSegcIMU{}, \LGKFgSegbIMU{} against the OMC system. 
        Only the results of the NR PiG and TCD dataset are presented, as the result of NR KFit are almost identical to NR PiG.
    Refer to code repository for links to video reconstruction of sample trials \cite{Sy-ckf2019-supp}.

\begin{table}[htbp]
    \centering
    \caption{Total travelled distance (TTD) deviation, stride length, and gait speed for \LGKFgSeg{} and optical motion capture (OMC) system}
    \label{tab:results-ttddev-sl-gs}%
    \begin{tabular}{cclrrrrrrrrr}
    \hline \hline
          &       & \parbox[t]{1mm}{\multirow{3}{*}{\rotatebox[origin=c]{90}{Side}}} & \multicolumn{1}{c}{TTD} & \multicolumn{4}{c}{Stride length (cm)} & \multicolumn{4}{c}{Gait speed (cm/s)} \\
          &       &       & \multicolumn{1}{c}{Error} & \multicolumn{2}{c}{Actual} & \multicolumn{2}{c}{Error} & \multicolumn{2}{c}{Actual} & \multicolumn{2}{c}{Error} \\
    \multicolumn{1}{l}{\begin{sideways}DB\end{sideways}} & Algo. &       & \multicolumn{1}{c}{\smallpct dev} & \multicolumn{1}{c}{$\mu$} & \multicolumn{1}{c}{med.} & \multicolumn{1}{c}{$\mu \smallpm \sigma$} & RMS   & \multicolumn{1}{c}{$\mu$} & \multicolumn{1}{c}{med.} & \multicolumn{1}{c}{$\mu \smallpm \sigma$} & RMS \\ \hline
    \multirow{8}[7]{*}{\begin{sideways}NeuRA\end{sideways}} & Freewalk & L     & 1.9\smallpct & 91    & 99    & $\smallneg1.7 \smallpm 5$ & 4.9   & 70    & 74    & $\smallneg1.2 \smallpm 4$ & 3.8 \\
          & \LGKFgSegcIMU{} & R     & 3.0\smallpct & 93    & 99    & $\smallneg2.8 \smallpm 5$ & 6.0   & 71    & 75    & $\smallneg1.9 \smallpm 4$ & 4.1 \bigstrut[b]\\
    \cline{2-12}      & Freewalk & L     & 3.7\smallpct & 91    & 99    & $\smallneg3.3 \smallpm 5$ & 5.9   & 70    & 74    & $\smallneg2.5 \smallpm 4$ & 4.7 \bigstrut[t]\\
          & \LGKFgSegbIMU{} & R     & 4.8\smallpct & 93    & 99    & $\smallneg4.4 \smallpm 6$ & 7.1   & 71    & 75    & $\smallneg3.3 \smallpm 4$ & 5.3 \bigstrut[b]\\
    \cline{2-12}      & Jog   & L     & 10.3\smallpct & 81    & 86    & $\smallneg8.4 \smallpm 15$ & 17.3  & 107   & 118   & $\smallneg9.2 \smallpm 24$ & 26.0 \bigstrut[t]\\
          & \LGKFgSegcIMU{} & R     & 11.8\smallpct & 85    & 97    & $\smallneg10.1 \smallpm 19$ & 21.2  & 111   & 124   & $\smallneg11.6 \smallpm 28$ & 29.8 \bigstrut[b]\\
    \cline{2-12}      & Jog   & L     & 9.9\smallpct & 81    & 86    & $\smallneg8.1 \smallpm 15$ & 17.2  & 107   & 118   & $\smallneg8.7 \smallpm 24$ & 25.5 \bigstrut[t]\\
          & \LGKFgSegbIMU{} & R     & 11.2\smallpct & 85    & 97    & $\smallneg9.5 \smallpm 18$ & 20.4  & 111   & 124   & $\smallneg11.1 \smallpm 27$ & 28.8 \bigstrut[b]\\
    \hline
    \multirow{4}[4]{*}{\begin{sideways}TCD\end{sideways}} & Walking & L     & 3.7\smallpct & 117   & 124   & $\smallneg4.4 \smallpm 5$ & 6.7   & 97    & 97    & $\smallneg3.4 \smallpm 4$ & 5.3 \bigstrut[t]\\
          & \LGKFgSegcIMU{} & R     & 4.3\smallpct & 115   & 121   & $\smallneg4.9 \smallpm 5$ & 7.2   & 96    & 96    & $\smallneg4.0 \smallpm 4$ & 5.8 \bigstrut[b]\\
    \cline{2-12}      & Walking & L     & 4.2\smallpct & 117   & 124   & $\smallneg5.0 \smallpm 5$ & 7.2   & 97    & 97    & $\smallneg3.9 \smallpm 5$ & 6.0 \bigstrut[t]\\
          & \LGKFgSegbIMU{} & R     & 4.3\smallpct & 115   & 121   & $\smallneg5.0 \smallpm 6$ & 7.5   & 96    & 96    & $\smallneg4.0 \smallpm 4$ & 6.0 \bigstrut[b]\\
    \hline
    \hline
    \end{tabular}%

   where $\mu$, med., and $\sigma$ denote mean, median, and standard deviation.
\end{table}%
    
\section{Discussion}

\subsection{Mean position and orientation RMSE}
The mean position and orientation RMSE gives a performance overview of the different algorithms.
    Both \LGKFgSegcIMU{} and \LGKFgSegbIMU{} are comparable to \emph{OSPS} (\texttildelow$\Delta 1$ cm, \texttildelow$\Delta 1.5^\circ$) for free walking.
        Note that the $e_{pos}$ of \LGKFeSegcIMU{} was calculated from six joints/points (hips to ankles), while the other algorithms were calculated from eight joint/points (hips to toes).
    \LGKFgSegcIMU{}'s $e_{pos}$ is ~$0.5$ cm better than \LGKFgSegbIMU{}, which is expected as \LGKFgSegcIMU{} utilises more IMU sensor units.
        Interestingly, \LGKFgSegcIMU{}'s $e_{ori}$ is $1^\circ$ worse than \LGKFgSegbIMU{}.
        This is probably due to the additional uninstrumented segment (\ie{} pelvis) for the two-IMU setup being tracked more accurately than the thigh and shank orientations.
    Lastly, the results for dynamic movements were worse, notably for the TCD dataset ($\Delta 7$ cm, $\Delta 11^\circ$, compared to free walking results),
        which was expected as the TCD dataset contains movements that break our pelvis pseudo-measurement assumptions (\eg{} during the crawling movement, the pelvis $z$ position is close to the floor instead of being at standing height; while the pelvis $x$ and $y$ position is no longer between the feet $x$ and $y$ positions).

Comparing the PiG and KFit model of the NR database, 
    $e_{pos}$ was consistent ($<\Delta 0.5 $ cm) for \LGKFgSegcIMU{}, \LGKFgSegbIMU{}, and \emph{OSPS}.
    However, the $e_{ori}$ of the KFit model was \texttildelow$4^\circ$ better than the PiG model,
        which is understandable given that the KFit model implements assumptions similar to our constraints (\eg{} constant body segment lengths and hinged knee joints).

\LGKFgSeg{} was also comparable to existing algorithms in the literature.
    Both $e_{pos}$ and $e_{ori}$ for NR PiG were similar to Sy \etal{}'s work \cite{syckf2020}  (\eg{} $5.3$ - $5.7$ cm and $14.2^\circ$ - $15.5^\circ$ compared to Sy's $5.2$ cm and $16.1^\circ$ for free walking).
    Our $e_{ori}$ for NR KFit was also similar to Marcard \etal{}'s sparse inertial poser (SIP) (\eg{} $10.6^\circ$-$10.9^\circ$ compared to SIP's \texttildelow$11^\circ$ for free walking, \texttildelow$13.3^\circ$ compared to SIP's \texttildelow$15^\circ$ for dynamic movements) \cite{Marcard2017}.
        However, there was a bigger gap for $e_{pos}$ (\eg{} \texttildelow$5.5$ cm against SIP's \texttildelow$3$ cm for free walking, \texttildelow$9.5$ cm against SIP's \texttildelow$5$ cm for dynamic movements).
    For the TCD dataset, our $e_{pos}$ result for free walking ($6.94$-$7.42$ cm) was close to Gilbert \etal{}'s neural network approach ($5.2$-$6.4$ cm) \cite{Gilbert2019}.
    Although we expected our algorithm to have a worse performance compared to SIP and Gilbert \etal{}'s, as \LGKFgSeg{} has significantly less computation cost \cite{Marcard2017, Gilbert2019}.
        The performance gap may also have widened due to the inherent difficulty of tracking more uninstrumented linked segments between the instrumented segments (\ie{} the furthest IMUs from the pelvis for the \LGKFgSeg{} were at the feet, while the said literature had the IMUs at the shanks).
    For reference on computational cost, \LGKFgSeg{} processed a 1,000-frame sequence in \texttildelow$2$ seconds on a Intel Core i5-6500 3.2 GHz CPU, while \emph{SIP} took 7.5 minutes on a quad-core Intel Core i7 3.5 GHz CPU. 
        Though some deep learning based approaches boast real time processing, they still require computing resources with graphic processing units (GPUs), limiting its feasibility in mobile applications.

\subsection{Joint angle RMSE and CC}
The joint angle RMSE and CC provides a more in-depth (per joint) analysis of our algorithm's performance.
    Similar to \LGKFeSegcIMU{}, both \LGKFgSeg{} algorithms had good CC in the sagittal plane ($>0.6$ CC), with less promising results in the frontal and transverse plane \cite{sylgcekf2020}.
For both free walking and dynamic movements, the joint angle RMSE of both \LGKFgSegcIMU{} and \LGKFgSegbIMU{} were comparable (\texttildelow$\Delta 1^\circ $).
During free walking, the ankle and knee Y joint angles CC of both algorithms were comparable ($\Delta0.05$ CC),
    while the \LGKFgSegbIMU{} hip joint angles CC, calculated from pelvis and thigh orientation, were slightly worse than \LGKFgSegcIMU{} ($\Delta 0.1$ CC),
    which is probably due to both pelvis and thigh having no sensors attached.
During dynamic movements, \LGKFgSegcIMU{}'s performance deteriorated (\eg{} $<0.3$ CC on NR PiG and KFit at the sagittal plane) compared to free walking movements.
    \LGKFgSegbIMU{} had an even worse performance ($<0.5$ CCs) especially for jumping jacks (\texttildelow$\Delta 0.5$ CC deterioration).
        
Comparing between the PiG and KFit model of the NR database for free walking, the ankle, knee, and hip Y angles were similar ($\Delta 0.1$ CC).
    The most notable difference was with the hip Z joint angle, which improved significantly by $5$ - $8^\circ$ RMSE and $0.3$ - $0.5$ CC (KFit model was better).
    The said dramatic improvement was probably due to the similarity of the constant body segment length and hinged knee joint assumptions used by our algorithm.
Since the TCD dataset used a similar model as NR KFit, it is no surprise that the results from TCD are more similar to NR KFit than NR PiG during free walking (\eg{} more similar hip Z joint angles).

Table \ref{tab:per-step-joint-angle-rmse-cc} shows the joint angle RMSE and CC for our algorithm and related literature \cite{syckf2020, Hu2015}.
    For a more similar comparison, we computed the joint angles of \LGKFgSeg{} under a similar setup to existing literature (\ie{} each step from the straight walking part of NR dataset's walk movement was considered as a trial).
    \LGKFgSegcIMU{} and \LGKFgSegbIMU{} were comparable to Hu \etal{} \cite{Hu2015} with the biggest difference in hip Y (\texttildelow$\Delta 0.1$ CC) which maybe due to \cite{Hu2015} attaching two IMUs to the pelvis (we only used one),
    or due to the additional ambiguity that comes with tracking human body pose in 3D instead of tracking in 2D.

\begin{table}[htbp]
  \centering
  \caption{Per-step joint angle RMSE and CC for straight walks}
    \begin{tabular}{cccccccc}
    \hline
    \hline
          &       & \multicolumn{3}{c}{RMSE (no bias) $^\circ$} & \multicolumn{3}{c}{CC} \bigstrut\\
    \hline
    \multicolumn{2}{c}{Algorithm} & Hip   & Knee  & Ankle & Hip   & Knee  & Ankle \bigstrut\\
    \hline
    \multicolumn{1}{c}{\multirow{2}[4]{*}{NR\newline{}PiG}} & \multicolumn{1}{l}{\LGKFgSegcIMU{}} & $5.0 \smallpm 1.0$ & $8.2 \smallpm 2.2$ & $5.9 \smallpm 1.6$ & $0.95 \smallpm 0.02$ & $0.91 \smallpm 0.06$ & $0.78 \smallpm 0.18$ \bigstrut\\
    \cline{2-8}      & \multicolumn{1}{l}{\LGKFgSegbIMU{}} & $6.9 \smallpm 1.3$ & $9.5 \smallpm 2.0$ & $7.5 \smallpm 1.1$ & $0.90 \smallpm 0.05$ & $0.89 \smallpm 0.05$ & $0.77 \smallpm 0.13$ \bigstrut\\
    \hline
    \multicolumn{1}{c}{\multirow{2}[4]{*}{NR\newline{}Kfit}} & \multicolumn{1}{l}{\LGKFgSegcIMU{}} & $5.1 \smallpm 1.0$ & $8.4 \smallpm 1.8$ & $5.6 \smallpm 1.6$ & $0.95 \smallpm 0.03$ & $0.93 \smallpm 0.05$ & $0.74 \smallpm 0.19$ \bigstrut\\
    \cline{2-8}      & \multicolumn{1}{l}{\LGKFgSegbIMU{}} & $7.3 \smallpm 1.4$ & $9.0 \smallpm 1.9$ & $6.7 \smallpm 1.1$ & $0.88 \smallpm 0.06$ & $0.92 \smallpm 0.04$ & $0.74 \smallpm 0.14$ \bigstrut\\
    \hline
    \multicolumn{2}{c}{CKF-3IMU \cite{syckf2020}} & $4.4 \smallpm 1.9$ & $5.7 \smallpm 2.2$ & --     & $0.96 \smallpm 0.08$ & $0.98 \smallpm 0.03$ & -- \bigstrut\\
    \hline
    \multicolumn{2}{c}{Hu \etal \cite{Hu2015}} & $6.8 \smallpm 3.0$ & $4.9 \smallpm 3.5$ & $7.3 \smallpm 3.5$ & $0.97 \smallpm 0.04$ & $0.95 \smallpm 0.04$ & $0.79 \smallpm 0.17$ \bigstrut\\
    \hline
    \multicolumn{2}{c}{Tong \etal \cite{Li2020}} & $6.1 \smallpm 2.7$ & $6.3 \smallpm 2.4$ & $6.7 \smallpm 2.8$ & $0.97 \smallpm 0.06$ & $0.84 \smallpm 0.14$ & $0.65 \smallpm 0.18$ \bigstrut\\
    \hline
    \hline
    \end{tabular}%
    \label{tab:per-step-joint-angle-rmse-cc}%
\end{table}%

\subsection{Spatiotemporal gait parameters}
Table \ref{tab:results-ttddev-sl-gs} shows that in addition to tracking joint kinematics, 
    \LGKFgSegcIMU{} and \LGKFgSegbIMU{} can also track spatiotemporal parameters for free walking well ($1.9$-$4.8$\% TTD deviation), 
    although not at the same level as state-of-the-art dead reckoning algorithms \cite{Jimenez2010a, Zhang2017a} ($0.2$ - $1.5$\% TTD deviation).
Expectedly, \LGKFgSegbIMU{} performed slightly worse than \LGKFgSegcIMU{} (\texttildelow$\Delta 1$\% TTD error) which is most likely due to dead reckoning error/drift in the pelvis.
    As information is propagated between the body segments through the measurement and constraint update of \LGKFgSeg{}, the error in the pelvis can propagate to the feet.
    
\subsection{Towards monitoring activities of daily living (ADL)}
We have shown that \LGKFgSeg{} was able to track the full lower body motion using only two or three IMUs,
    notably achieving good joint angle CCs in the sagittal plane ($0.6$-$0.9$ CCs for free walking).
    However, the accuracy will need to improve (joint angle RMSE $<5^\circ$) to achieve clinical utility (usually involves walking movements), as well as to successfully track more ADL including dynamic movements.
To achieve better performance, one may leverage long-term recordings by averaging out cycle-to-cycle variation in estimation errors over many gait cycles, 
    or use additional sensing modalities, preferably packaged such that the number of sensor units will not increase.
    For example, distance ranging measurements or pressure insoles can be used to infer position of the pelvis \cite{syckfdist2020, Refai2020}.
    Attaching cameras to the body is another interesting approach (\eg{} Xu \etal{} tracked body pose using cap-mounted fisheye cameras pointing downwards \cite{Xu2019}; using cameras with IMUs for better position estimation \cite{Hol2008}).
    The extent to which these possible solutions can bridge the gap to clinical application and the tracking of ADLs and dynamic movements remains to be seen.
    
Additional considerations regarding the measurement and constraint assumptions used by our \LGKFgSeg{} algorithms must be given before use in everyday life.
    The pelvis pseudo-measurements and ankle flat-floor assumptions prevent accurate tracking of non-walking movements (\eg{} sleeping, crawling, high kicks).
    \LGKFgSeg{} will not be able to measure gait parameters where pathologies are present that break the hinged knee and ankle joint assumptions.
    Indeed, Kainz \etal{} recommends the use of models with fewer degrees-of-freedom (DoF) (\eg{} \LGKFgSeg{}) when studying healthy individuals due to better reliability,
        but to use models with more DoF (\eg{} 3-2-2 DoF of hip, knee, and ankle) when studying individuals with pathology \cite{Kainz2017a}.

Accurate step detection and sensor-to-body calibration algorithms will be needed to move towards a full remote gait monitoring system.
    Similar to \CKFcIMU{}, \LGKFgSeg{} relies heavily on accurate step detection and sensor-to-body calibration.
    Since the foot/shoe is already instrumented with an IMU, adding in-shoe pressure sensor to improve step detection accuracy will most likely not affect the user's comfort.
    Assuming the foot's frame coincides with the shoe, and the IMU is rigidly attached to the shoe, the sensor-to-shoe rotation offset can be ensured by design.
        However, a practical initial or even online calibration procedure will be needed for the pelvis sensor-to-body calibration (for \LGKFgSeg{}), and to align with the reference frames of the other IMUs.
        This calibration can be done through manual alignment from palpation of anatomical landmarks, use of an external calibration device \cite{Picerno2008}, or the subject may be asked to walk in a straight line and then back to the starting point for yaw offset alignment.
        
Lastly, although \LGKFgSegbIMU{}, which only uses two IMUs, is unable to accurately track non-walking movements (\eg{} crawling) due to the lack of sensor at the pelvis, it enjoys better numerical stability and easier sensor-to-body calibration because the algorithm is only \changed{optimizing the best lower body pose estimate from} two body segments, instead of three.

\section{Conclusion}
This paper described a Lie group constrained extended Kalman filter based algorithm that tracks the full lower body (seven body segments) using only two or three IMUs.
    The algorithm was extensively evaluated on two public datasets showing its performance compared to two standard benchmark approaches (\ie{} plug-in gait commonly used in gait analysis and kinematic fit commonly used in animation, robotics, and muscolo-skeleta simulation),
    giving insight into the similarity and differences between the said approaches used in different application areas.
The overall mean body segment position (relative to mid-pelvis origin) and orientation error of our algorithm for free walking was $5.93 \pm 1.33$ cm and $13.43 \pm 1.89^\circ$ when using three IMUs, 
    and $6.35 \pm 1.20$ cm and $12.71 \pm 1.60^\circ$ when using two IMUs.
The algorithm was able to track the joint angles in the sagittal plane for straight walking well, but requires improvement for unscripted movements (\eg{} turning around, side steps), especially for dynamic movements or when considering clinical applications.
    Nevertheless, this work has brought us closer to remote gait monitoring even when only using IMUs on the shoes.
    The low computation cost also signifies that it can be used with gait assistive devices.
Lastly, the source code will be made available at \githubLGKFgS{}.

\section{Acknowledgement}
This research was supported by an Australian Government Research Training Program Scholarship.

\section*{References}
\printbibliography[heading=none]

@book{Chirikjian2012Book2,
author = {Chirikjian, Gregory},
booktitle = {Stochastic Models, Information Theory, and Lie Groups},
doi = {10.1007/978-0-8176-4944-9},
title = {{Stochastic models, information theory, and Lie groups. II: Analytic methods and modern applications}},
year = {2012}
}

@article{Jimenez2010a,
author = {Jimenez, A. R. and Seco, F. and Prieto, J. C. and Guevara, J.},
doi = {10.1109/WPNC.2010.5649300},
isbn = {9781424471577},
journal = {Proceedings of the 2010 7th Workshop on Positioning, Navigation and Communication, WPNC'10},
number = {April},
pages = {135--143},
title = {{Indoor Pedestrian navigation using an INS/EKF framework for yaw drift reduction and a foot-mounted IMU}},
year = {2010}
}

@article{Delp2007a,
author = {Delp, Scott L. and Anderson, Frank C. and Arnold, Allison S. and Loan, Peter and Habib, Ayman and John, Chand T. and Guendelman, Eran and Thelen, Darryl G.},
doi = {10.1109/TBME.2007.901024},
issn = {00189294},
journal = {IEEE Transactions on Biomedical Engineering},
month = {nov},
number = {11},
pages = {1940--1950},
pmid = {18018689},
title = {{OpenSim: Open-source software to create and analyze dynamic simulations of movement}},
volume = {54},
year = {2007}
}

@article{VonMarcard2016,
author = {{Von Marcard}, Timo and Pons-Moll, Gerard and Rosenhahn, Bodo},
doi = {10.1109/TPAMI.2016.2522398},
issn = {01628828},
journal = {IEEE Transactions on Pattern Analysis and Machine Intelligence},
number = {8},
pages = {1533--1547},
title = {{Human Pose Estimation from Video and IMUs}},
volume = {38},
year = {2016}
}

@article{Zhang2017a,
author = {Zhang, Wenchao and Li, Xianghong and Wei, Dongyan and Ji, Xinchun and Yuan, Hong},
doi = {10.1109/IPIN.2017.8115916},
isbn = {9781509062980},
journal = {2017 International Conference on Indoor Positioning and Indoor Navigation, IPIN 2017},
number = {September},
pages = {1--5},
title = {{A foot-mounted PDR System Based on IMU/EKF+HMM+ZUPT+ZARU+HDR+compass algorithm}},
volume = {2017-Janua},
year = {2017}
}

@inproceedings{Marcard2017,
archivePrefix = {arXiv},
arxivId = {1703.08014},
author = {von Marcard, Timo and Rosenhahn, Bodo and Black, Michael J and Pons-Moll, Gerard},
booktitle = {Computer Graphics Forum},
doi = {10.1111/cgf.13131},
eprint = {1703.08014},
issn = {14678659},
number = {2},
organization = {Wiley Online Library},
pages = {349--360},
title = {{Sparse inertial poser: Automatic 3D human pose estimation from sparse IMUs}},
url = {https://ps.is.tuebingen.mpg.de/uploads{\_}file/attachment/attachment/345/sparseInertialPoser.pdf https://arxiv.org/pdf/1703.08014v2.pdf{\%}0Ahttps://ps.is.tuebingen.mpg.de/uploads{\_}file/attachment/attachment/345/sparseInertialPoser.pdf},
volume = {36},
year = {2017}
}

@inproceedings{Hu2015,
author = {Hu, Xinyao and Yao, Cheng and Soh, Gim Song},
booktitle = {IEEE International Conference on Rehabilitation Robotics},
doi = {10.1109/ICORR.2015.7281257},
isbn = {9781479918072},
issn = {19457901},
pages = {549--554},
title = {{Performance evaluation of lower limb ambulatory measurement using reduced inertial measurement units and 3R gait model}},
year = {2015}
}

@misc{Sy-ckf2019-supp,
title = {gait-tech/gt.papers},
url = {https://github.com/gait-tech/gt.papers},
urldate = {2019-12-17}
}

@article{syckfdist2020,
archivePrefix = {arXiv},
arxivId = {2003.10228},
author = {Sy, Luke and Lovell, Nigel H. and Redmond, Stephen J.},
eprint = {2003.10228},
journal = {2020 42nd Annual International Conference of the IEEE Engineering in Medicine and Biology Society (EMBC)},
month = {mar},
title = {{Estimating lower limb kinematics using distance measurements with a reduced wearable inertial sensor count}},
url = {http://arxiv.org/abs/2003.10228},
year = {2020}
}

@article{Barfoot2014,
author = {Barfoot, Timothy D and Furgale, Paul T},
doi = {10.1109/TRO.2014.2298059},
issn = {15523098},
journal = {IEEE Transactions on Robotics},
number = {3},
pages = {679--693},
publisher = {IEEE},
title = {{Associating uncertainty with three-dimensional poses for use in estimation problems}},
volume = {30},
year = {2014}
}

@article{Gilbert2019,
author = {Gilbert, Andrew and Trumble, Matthew and Malleson, Charles and Hilton, Adrian and Collomosse, John},
doi = {10.1007/s11263-018-1118-y},
issn = {15731405},
journal = {International Journal of Computer Vision},
number = {4},
pages = {381--397},
publisher = {Springer US},
title = {{Fusing visual and inertial sensors with semantics for 3D human pose estimation}},
url = {https://doi.org/10.1007/s11263-018-1118-y},
volume = {127},
year = {2019}
}

@article{Wouda2019,
author = {Wouda, Frank J. and Giuberti, Matteo and Rudigkeit, Nina and van Beijnum, Bert-Jan F. and Poel, Mannes and Veltink, Peter H.},
doi = {10.3390/s19173716},
issn = {1424-8220},
journal = {Sensors},
number = {17},
pages = {3716},
title = {{Time Coherent Full-Body Poses Estimated Using Only Five Inertial Sensors: Deep versus Shallow Learning}},
url = {https://www.mdpi.com/1424-8220/19/17/3716},
volume = {19},
year = {2019}
}

@inproceedings{Vlasic2007,
address = {New York, New York, USA},
author = {Vlasic, Daniel and Adelsberger, Rolf and Vannucci, Giovanni and Barnwell, John and Gross, Markus and Matusik, Wojciech and Popovi{\'{c}}, Jovan},
booktitle = {ACM Transactions on Graphics},
doi = {10.1145/1276377.1276421},
isbn = {07300301},
issn = {07300301},
number = {3},
pages = {35},
publisher = {ACM Press},
title = {{Practical motion capture in everyday surroundings}},
url = {http://portal.acm.org/citation.cfm?doid=1276377.1276421},
volume = {26},
year = {2007}
}

@article{Wouda2016,
author = {Wouda, Frank F.J. and Giuberti, Matteo and Bellusci, Giovanni and Veltink, Peter P.H.},
doi = {10.3390/s16122138},
issn = {1424-8220},
journal = {Sensors},
month = {dec},
number = {12},
pages = {2138},
publisher = {Multidisciplinary Digital Publishing Institute},
title = {{Estimation of full-body poses using only five inertial sensors: an eager or lazy learning approach?}},
url = {http://www.mdpi.com/1424-8220/16/12/2138},
volume = {16},
year = {2016}
}

@inproceedings{Trumble2017,
author = {Trumble, Matthew and Gilbert, Andrew and Hilton, Adrian and Collomosse, John},
booktitle = {British Machine Vision Conference (BMVC)},
isbn = {0042-6822 (Print)$\backslash$r0042-6822 (Linking)},
number = {September},
pages = {1--13},
pmid = {3617505},
title = {{Total Capture: 3D Human Pose Estimation Fusing Video and Inertial Sensors}},
url = {http://cvssp.org/data/totalcapture/.},
year = {2017}
}

@article{DelRosario2018,
author = {{Del Rosario}, Michael B. and Khamis, Heba and Ngo, Phillip and Lovell, Nigel H. and Redmond, Stephen J.},
doi = {10.1109/JSEN.2018.2864989},
issn = {1530437X},
journal = {IEEE Sensors Journal},
number = {22},
pages = {9332--9342},
publisher = {IEEE},
title = {{Computationally efficient adaptive error-state Kalman filter for attitude estimation}},
url = {https://ieeexplore.ieee.org/document/8434202/},
volume = {18},
year = {2018}
}

@inproceedings{Cloete2008,
author = {Cloete, Teunis and Scheffer, Cornie},
booktitle = {2008 30th Annual International Conference of the IEEE Engineering in Medicine and Biology Society},
doi = {10.1109/IEMBS.2008.4650232},
isbn = {978-1-4244-1814-5},
month = {aug},
pages = {4579--4582},
publisher = {IEEE},
title = {{Benchmarking of a full-body inertial motion capture system for clinical gait analysis}},
url = {http://ieeexplore.ieee.org/document/4650232/},
year = {2008}
}

@article{DelRosario2016a,
author = {{Del Rosario}, Michael B. and Lovell, Nigel H. and Redmond, Stephen J.},
doi = {10.1109/JSEN.2016.2574124},
issn = {1530437X},
journal = {IEEE Sensors Journal},
number = {15},
pages = {6008--6017},
title = {{Quaternion-based complementary filter for attitude determination of a smartphone}},
url = {https://www.researchgate.net/profile/Michael{\_}Del{\_}Rosario/publication/303598655{\_}Quaternion-Based{\_}Complementary{\_}Filter{\_}for{\_}Attitude{\_}Determination{\_}of{\_}a{\_}Smartphone/links/57ceabbb08aed6789700e24e.pdf},
volume = {16},
year = {2016}
}

@article{Kainz2017a,
author = {Kainz, Hans and Graham, David and Edwards, Julie and Walsh, Henry P.J. and Maine, Sheanna and Boyd, Roslyn N. and Lloyd, David G. and Modenese, Luca and Carty, Christopher P.},
doi = {10.1016/j.gaitpost.2017.04.001},
issn = {18792219},
journal = {Gait and Posture},
number = {April},
pages = {325--331},
pmid = {28411552},
publisher = {Elsevier},
title = {{Reliability of four models for clinical gait analysis}},
url = {http://dx.doi.org/10.1016/j.gaitpost.2017.04.001},
volume = {54},
year = {2017}
}

@article{sylgkf5seg2020,
archivePrefix = {arXiv},
arxivId = {1910.01808},
author = {Sy, Luke Wicent F and Lovell, Nigel H and Redmond, Stephen J},
doi = {10.3390/s20236829},
eprint = {1910.01808},
isbn = {9781728159072},
issn = {1424-8220},
journal = {Sensors},
number = {23},
pages = {310--315},
title = {{Estimating Lower Limb Kinematics Using a Lie Group Constrained Extended Kalman Filter with a Reduced Wearable IMU Count and Distance Measurements}},
url = {https://www.mdpi.com/1424-8220/20/23/6829},
volume = {20},
year = {2020}
}

@article{Roetenberg2009,
author = {Roetenberg, Daniel and Luinge, Henk and Slycke, Per},
doi = {10.1.1.569.9604},
isbn = {9781595936349},
journal = {Xsens Motion Technologies BV, Tech. Rep},
title = {{Xsens MVN: Full 6DOF human motion tracking using miniature inertial sensors}},
url = {https://www.xsens.com/images/stories/PDF/MVN{\_}white{\_}paper.pdf},
volume = {1},
year = {2009}
}

@article{Fakult2013,
author = {Fakult, Der and Wilhelm, Gottfried and Universit, Leibniz},
pages = {1--14},
title = {{Human pose estimation from video and inertial sensors}},
year = {2013}
}

@article{Salarian2013,
author = {Salarian, Arash and Burkhard, Pierre R. and Vingerhoets, Fran{\c{c}}ois J. G. and Jolles, Brigitte M. and Aminian, Kamiar},
doi = {10.1109/TBME.2012.2223465},
issn = {0018-9294},
journal = {IEEE Transactions on Biomedical Engineering},
month = {jan},
number = {1},
pages = {72--77},
title = {{A novel approach to reducing number of sensing units for wearable gait analysis systems}},
url = {http://ieeexplore.ieee.org/document/6327615/},
volume = {60},
year = {2013}
}

@article{Refai2020,
author = {Refai, Mohamed Irfan Mohamed and {Van Beijnum}, Bert Jan F. and Buurke, Jaap H. and Veltink, Peter H.},
doi = {10.1109/TNSRE.2020.2984809},
issn = {15580210},
journal = {IEEE Transactions on Neural Systems and Rehabilitation Engineering},
number = {6},
pages = {1308--1316},
pmid = {32310775},
title = {{Portable Gait Lab: Estimating 3D GRF Using a Pelvis IMU in a Foot IMU Defined Frame}},
volume = {28},
year = {2020}
}

@article{Joukov2017,
author = {Joukov, Vladimir and Cesic, Josip and Westermann, Kevin and Markovic, Ivan and Kulic, Dana and Petrovic, Ivan},
doi = {10.1109/IROS.2017.8206016},
isbn = {9781538626825},
issn = {21530866},
journal = {IEEE International Conference on Intelligent Robots and Systems},
month = {dec},
pages = {1965--1972},
publisher = {Institute of Electrical and Electronics Engineers Inc.},
title = {{Human motion estimation on Lie groups using IMU measurements}},
volume = {2017-Septe},
year = {2017}
}

@article{Li2020,
author = {Li, Tong and Wang, Lei and Li, Qingguo and Liu, Tao},
doi = {10.1109/AIM43001.2020.9158961},
isbn = {9781728167947},
journal = {IEEE/ASME International Conference on Advanced Intelligent Mechatronics, AIM},
pages = {1143--1148},
title = {{Lower-body walking motion estimation using only two shank-mounted inertial measurement units (IMUs)}},
volume = {2020-July},
year = {2020}
}

@book{stillwell2008naive,
author = {Stillwell, John},
publisher = {Springer Science {\&} Business Media},
title = {{Naive Lie theory}},
year = {2008}
}

@article{Tautges2011,
archivePrefix = {arXiv},
arxivId = {1006.4903},
author = {Tautges, Jochen and Zinke, A. and Kr{\"{u}}ger, B and Baumann, J and Weber, Andreas and Helten, T and M{\"{u}}ller, M and Seidel, H.P. and Eberhardt, B.},
doi = {10.1145/PREPRINT},
eprint = {1006.4903},
isbn = {0103660054},
issn = {07300301},
journal = {ACM Transactions on Graphics (TOG)},
number = {3},
pages = {18},
title = {{Motion reconstruction using sparse accelerometer data}},
url = {http://doi.acm.org/10.1145/1966394.1966397 http://portal.acm.org/citation.cfm?id=1966397},
volume = {30},
year = {2011}
}

@article{Cesic2016,
author = {{\'{C}}esi{\'{c}}, Josip and Joukov, Vladimir and Petrovi{\'{c}}, Ivan and Kuli{\'{c}}, Dana},
doi = {10.1109/HUMANOIDS.2016.7803369},
isbn = {9781509047185},
issn = {21640580},
journal = {IEEE-RAS International Conference on Humanoid Robots},
pages = {826--833},
title = {{Full body human motion estimation on lie groups using 3D marker position measurements}},
year = {2016}
}

@inproceedings{Huang2018a,
archivePrefix = {arXiv},
arxivId = {1810.04703},
author = {Huang, Yinghao and Kaufmann, Manuel and Aksan, Emre and Black, Michael J. and Hilliges, Otmar and Pons-Moll, Gerard},
booktitle = {SIGGRAPH Asia 2018 Technical Papers, SIGGRAPH Asia 2018},
doi = {10.1145/3272127.3275108},
eprint = {1810.04703},
isbn = {9781450360081},
issn = {15577368},
month = {dec},
publisher = {Association for Computing Machinery, Inc},
title = {{Deep inertial poser: Learning to reconstruct human pose from sparse inertial measurements in real time}},
year = {2018}
}

@article{syckf2020,
archivePrefix = {arXiv},
arxivId = {cs.RO/1910.00910},
author = {Sy, Luke Wicent and Raitor, Michael and {Del Rosario}, Michael B. and Khamis, Heba and Kark, Lauren and Lovell, Nigel Hamilton and Redmond, Stephen J.},
eprint = {1910.00910},
issn = {0018-9294},
journal = {IEEE Transactions on Biomedical Engineering},
month = {sep},
pages = {1--10},
primaryClass = {cs.RO},
publisher = {Institute of Electrical and Electronics Engineers (IEEE)},
title = {{Estimating lower limb kinematics using a reduced wearable sensor count}},
url = {https://arxiv.org/pdf/1910.00910.pdf https://ieeexplore.ieee.org/document/9205648/},
year = {2020}
}

@article{Le,
author = {Le, Quoc and Mikolov, Tomas},
title = {{Distributed Representations of Sentences and Documents}},
url = {http://proceedings.mlr.press/v32/le14.pdf}
}

@article{Wang2006,
author = {Wang, Yunfeng and Chirikjian, Gregory S.},
doi = {10.1109/TRO.2006.878978},
issn = {15523098},
journal = {IEEE Transactions on Robotics},
number = {4},
pages = {591--602},
title = {{Error propagation on the Euclidean group with applications to manipulator kinematics}},
volume = {22},
year = {2006}
}

@article{Hol2008,
author = {Hol, Jeroen D. and Sch{\"{o}}n, Thomas B. and Gustafsson, Fredrik},
doi = {10.1109/ISMAR.2008.4637318},
isbn = {9781424428403},
journal = {Proceedings - 7th IEEE International Symposium on Mixed and Augmented Reality 2008, ISMAR 2008},
pages = {21--24},
title = {{Relative pose calibration of a spherical camera and an IMU}},
year = {2008}
}

@incollection{selig2004lie,
author = {Selig, Jon M},
booktitle = {Computational Noncommutative Algebra and Applications},
pages = {101--125},
publisher = {Springer},
title = {{Lie groups and Lie algebras in robotics}},
year = {2004}
}

@article{Xu2019,
author = {Xu, Weipeng and Chatterjee, Avishek and Zollh, Michael and Rhodin, Helge and Fua, Pascal and Seidel, Hans-peter and Theobalt, Christian},
journal = {IEEE Transactions on Visualization and Computer Graphics (Proc. IEEE VR, 2019)},
number = {5},
pages = {2093--2101},
title = {{Mo2Cap2 : Real-time mobile 3D motion capture with a cap-mounted fisheye camera}},
volume = {25},
year = {2019}
}

@article{Picerno2008,
author = {Picerno, Pietro and Cereatti, Andrea and Cappozzo, Aurelio},
doi = {10.1016/j.gaitpost.2008.04.003},
isbn = {0966-6362 (Print)$\backslash$n0966-6362 (Linking)},
issn = {09666362},
journal = {Gait and Posture},
number = {4},
pages = {588--595},
pmid = {18502130},
title = {{Joint kinematics estimate using wearable inertial and magnetic sensing modules}},
url = {http://www.gaitposture.com/article/S0966-6362(08)00100-8/pdf http://ac.els-cdn.com/S0966636208001008/1-s2.0-S0966636208001008-main.pdf?{\_}tid=f459052a-9e6e-11e7-a430-00000aacb361{\&}acdnat=1505958684{\_}59e515a7a900874af96b35203e8225d8},
volume = {28},
year = {2008}
}

@inproceedings{Shull2010,
author = {Shull, Pete and Lurie, Kristen and Shin, Mihye and Besier, Thor and Cutkosky, Mark},
booktitle = {2010 IEEE Haptics Symposium},
doi = {10.1109/HAPTIC.2010.5444625},
isbn = {9781424468218},
issn = {2324-7347},
organization = {IEEE},
pages = {409--416},
title = {{Haptic gait retraining for knee osteoarthritis treatment}},
year = {2010}
}

@article{Kadaba1990,
author = {Kadaba, M. P. and Ramakrishnan, H. K. and Wootten, M. E.},
doi = {10.1002/jor.1100080310},
issn = {0736-0266},
journal = {Journal of Orthopaedic Research},
month = {may},
number = {3},
pages = {383--392},
publisher = {Raven Press, Ltd},
title = {{Measurement of Lower Extremity Kinematics During Level Walking}},
url = {http://doi.wiley.com/10.1002/jor.1100080310 http://www.biomech.uottawa.ca/english/teaching/apa6905/lectures/2012/Kadaba (1990) J Orthop Res.pdf},
volume = {8383},
year = {1990}
}

@inproceedings{sylgcekf2020,
archivePrefix = {arXiv},
arxivId = {1910.01808v1},
author = {Sy, Luke and Lovell, Nigel H and Redmond, Stephen J},
booktitle = {2020 8th IEEE International Conference on Biomedical Robotics and Biomechatronics (Biorob)},
eprint = {1910.01808v1},
title = {{Estimating lower limb kinematics using a Lie group constrained EKF and a reduced wearable IMU count}},
url = {https://github.com/lsy3/lg-cekf.},
year = {2020}
}

@article{Llorens2015,
author = {Llor{\'{e}}ns, Roberto and Gil-G{\'{o}}mez, Jos{\'{e}} Antonio and Alca{\~{n}}iz, Mariano and Colomer, Carolina and No{\'{e}}, Enrique},
doi = {10.1177/0269215514543333},
isbn = {1477-0873 (Electronic)$\backslash$r0269-2155 (Linking)},
issn = {14770873},
journal = {Clinical Rehabilitation},
number = {3},
pages = {261--268},
pmid = {25056999},
title = {{Improvement in balance using a virtual reality-based stepping exercise: A randomized controlled trial involving individuals with chronic stroke}},
volume = {29},
year = {2015}
}

@article{Leardini2017,
author = {Leardini, Alberto and Belvedere, Claudio and Nardini, Fabrizio and Sancisi, Nicola and Conconi, Michele and Parenti-Castelli, Vincenzo},
doi = {10.1016/j.jbiomech.2017.04.029},
issn = {18732380},
journal = {Journal of Biomechanics},
pages = {77--86},
publisher = {Elsevier Ltd},
title = {{Kinematic models of lower limb joints for musculo-skeletal modelling and optimization in gait analysis}},
url = {https://doi.org/10.1016/j.jbiomech.2017.04.029},
volume = {62},
year = {2017}
}

@inproceedings{Bourmaud2013,
author = {Bourmaud, Guillaume and Megret, Remi and Giremus, Audrey and Berthoumieu, Yannick},
booktitle = {European Signal Processing Conference},
isbn = {9780992862602},
issn = {22195491},
pages = {1--5},
title = {{Discrete extended Kalman filter on Lie groups}},
year = {2013}
}

@book{barfoot2017state,
author = {Barfoot, Timothy D},
publisher = {Cambridge University Press},
title = {{State Estimation for Robotics}},
year = {2017}
}

@article{Leardini2005,
author = {Leardini, Alberto and Chiari, Alberto and {Della Croce}, Ugo and Cappozzo, Aurelio and Chiari, Lorenzo and Croce, Ugo Della and Cappozzo, Aurelio},
doi = {10.1016/j.gaitpost.2004.05.002},
isbn = {0966-6362 (Print)$\backslash$n0966-6362 (Linking)},
issn = {0966-6362},
journal = {Gait and Posture},
month = {feb},
number = {2},
pages = {212--225},
pmid = {15639399},
publisher = {Elsevier},
title = {{Human movement analysis using stereophotogrammetry Part 3. Soft tissue artifact assessment and compensation}},
url = {http://www.sciencedirect.com/science/article/pii/S0966636204000773 http://ac.els-cdn.com/S0966636204000773/1-s2.0-S0966636204000773-main.pdf?{\_}tid=51df4716-9dbd-11e7-abf0-00000aab0f26{\&}acdnat=1505882390{\_}eef6cc90a5fe1169cc48c00d90dc8dc9},
volume = {21},
year = {2005}
}

@misc{sylgkf7segsupp2020,
title = {gait-tech/gt.papers/+lgkf7},
url = {https://github.com/gait-tech/gt.papers}
}

@article{VanDenNoort2013,
author = {{Van Den Noort}, Josien C. and Ferrari, Alberto and Cutti, Andrea G. and Becher, Jules G. and Harlaar, Jaap},
doi = {10.1007/s11517-012-1006-5},
isbn = {1151701210065},
issn = {01400118},
journal = {Medical and Biological Engineering and Computing},
number = {4},
pages = {377--386},
title = {{Gait analysis in children with cerebral palsy via inertial and magnetic sensors}},
volume = {51},
year = {2013}
}

\end{document}